\documentclass[lettersize,journal]{IEEEtran}
\usepackage{amsmath,amsfonts}
\usepackage{color}
\usepackage{array}
\usepackage[caption=false,font=normalsize,labelfont=sf,textfont=sf]{subfig}
\usepackage{textcomp}
\usepackage{caption}
\usepackage{booktabs}
\usepackage{multirow}
\usepackage{xspace}
\usepackage{stfloats}
\usepackage{url}
\usepackage{verbatim}
\usepackage{graphicx}
\usepackage{adjustbox}
\usepackage{cite}
\usepackage{bbding}
\usepackage{algorithm}
\usepackage{algpseudocode}
\usepackage{verbatim}

\begin{document}
\title{Geometric-Aware Low-Light Image and Video Enhancement via Depth Guidance}
\author{Yingqi Lin, Xiaogang Xu, Jiafei Wu, Yan Han and Zhe Liu
\IEEEcompsocitemizethanks{\IEEEcompsocthanksitem Yingqi Lin, Jiafei Wu, Yan Han and Zhe Liu work for Zhejiang Lab.

E-mail: \{linyq, hanyan, wujiafei, zhe.liu\}@zhejianglab.com
\IEEEcompsocthanksitem Xiaogang Xu is a ZJU100 Professor in Zhejiang University and a research scientist in Zhejiang Lab. This work is done when he was visiting Max Planck Institute for Informatics (Visual Computing and AI Department) at Saarland Informatics Campus as a postdoctoral research fellow.

E-mail: xiaogangxu00@gmail.com
\IEEEcompsocthanksitem Yingqi Lin and Xiaogang Xu contributed equally to this work. They are co-first authors
\IEEEcompsocthanksitem \textbf{\Envelope}Corresponding Author:  Xiaogang Xu \protect\\}}

\markboth{under review}
{Shell \MakeLowercase{\textit{et al.}}: A Sample Article Using IEEEtran.cls for IEEE Journals}

\maketitle

\begin{figure*}[tb]
\begin{center}
    \captionsetup{type=figure}
    \includegraphics[width=1\textwidth, height=6.2cm]{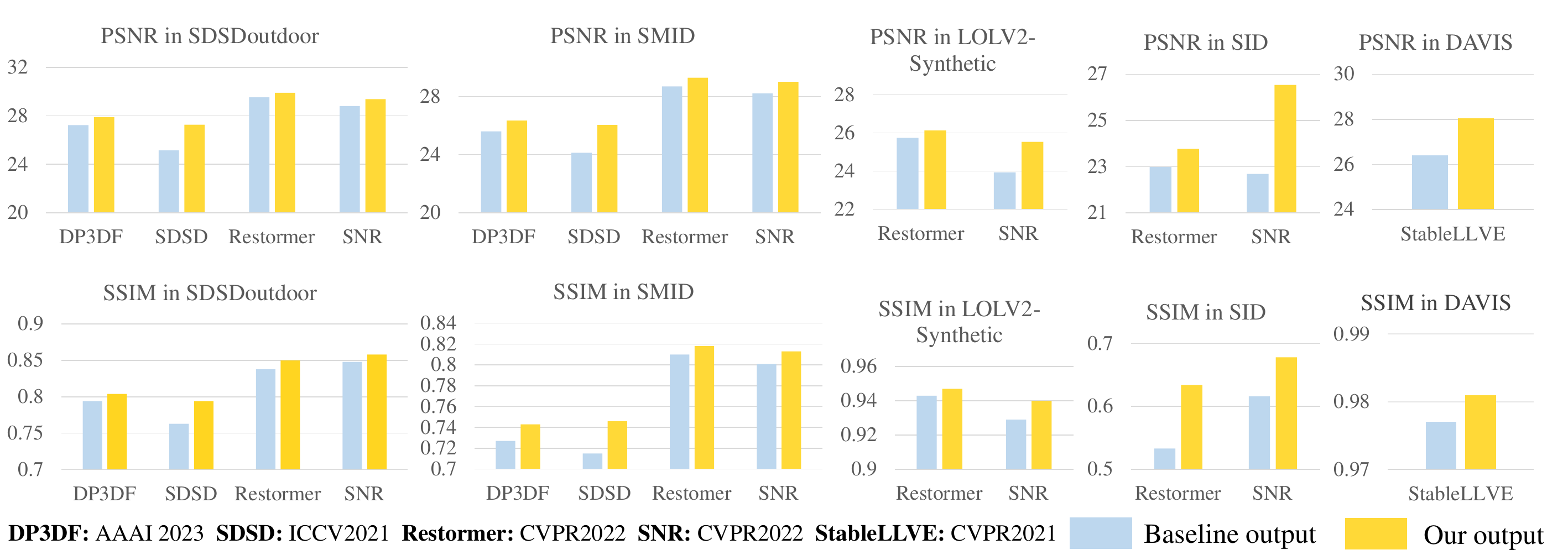}
    \captionof{figure}{We propose to enhance the performance of existing LLIE and LLVE methods via depth priors with our designed GG-LLERF. Our method consistently achieves the improvement target on different datasets/networks with the same structure to extract depth-aware features and complete the fusion with original images/frames' features. }
    \label{fig:teaser}
    \vspace{-0.1in}
\end{center}
\end{figure*}

\begin{abstract}
  Low-Light Enhancement (LLE) is aimed at improving the quality of photos/videos captured under low-light conditions. It is worth noting that most existing LLE methods do not take advantage of geometric modeling. We believe that incorporating geometric information can enhance LLE performance, as it provides insights into the physical structure of the scene that influences illumination conditions. To address this, we propose a Geometry-Guided Low-Light Enhancement Refine Framework (GG-LLERF) designed to assist low-light enhancement models in learning improved features by integrating geometric priors into the feature representation space. In this paper, we employ depth priors as the geometric representation. Our approach focuses on the integration of depth priors into various LLE frameworks using a unified methodology. This methodology comprises two key novel modules. First, a depth-aware feature extraction module is designed to inject depth priors into the image representation. Then, the Hierarchical Depth-Guided Feature Fusion Module (HDGFFM) is formulated with a cross-domain attention mechanism, which combines depth-aware features with the original image features within LLE models. We conducted extensive experiments on public low-light image and video enhancement benchmarks. The results illustrate that our framework significantly enhances existing LLE methods. 
  The source code and pre-trained models are available at \url{https://github.com/Estheryingqi/GG-LLERF}.
\end{abstract}

\textbf{}
\section{Introduction}
\label{sec:intro}

\IEEEPARstart{L}{ow-light} imaging is a common requirement in our daily lives, but it often results in poor-quality images or videos due to inadequate illumination or limited exposure time~\cite{chen2019seeing}. In response to this issue, the Low-Light Enhancement (LLE) technique has been developed. Its objectives include reducing noise and artifacts, preserving edges and textures, and reproducing natural brightness and color~\cite{wang2023lighting,yang2023implicit}. 

LLE can be further categorized into two subdomains: Low-Light Image Enhancement (LLIE)~\cite{xu2022snr,fu2023you,fu2023learning} and Low-Light Video Enhancement (LLVE)~\cite{wang2021sdsd,xu2023deep,zhang2021learning}. Both LLIE and LLVE can significantly benefit the performance of various downstream visual applications, such as nighttime detection and autopilot systems~\cite{lamba2020towards,wang2022sfnet}. Although many deep-learning-based LLE methods have been introduced and have achieved considerable success in some scenarios, their performance in more challenging situations is often unsatisfactory and has encountered certain limitations. Overcoming these performance limits is currently a prominent research topic.

One approach to improving enhancement performance involves the utilization of multi-modal maps as priors. For example, SKF~\cite{wu2023skf} leverages semantic maps to optimize the feature space for low-light enhancement; SMG~\cite{xu2023low} employs a generative framework to integrate edge information, thereby enhancing the initial appearance modeling for low-light scenarios. However, it's worth noting that all of these employed priors operate at the 2D level. 2D-level information may not fully capture the geometric information of the entire 3D scene. The inclusion of 3D geometric information can be instrumental in determining the illumination distribution in the scene, thereby enhancing LLE tasks. This is especially important in video tasks. Having geometric information of each frame allows for a more accurate modeling of the scene's geometric features.

In this paper, we introduce a groundbreaking approach by learning the essential geometric priors within Low-Light Enhancement frameworks. Subsequently, we develop an effective Geometry-Guided Low-Light Enhancement Refine Framework (GG-LLERF) tailored to incorporate these priors into the target frameworks $\mathcal{G}$, thereby surpassing their original performance limits. For our geometric representation, we harness depth priors, which offer valuable insights into the scene's geometry. We choose depth priors because of the existence of highly effective open-world depth estimation models $\mathcal{D}$ that have been trained on large-scale datasets, e.g., DPT~\cite{ranftl2021vision}.

Formulating a desired depth prior presents two primary challenges. First, achieving precise depth estimation directly from a low-light image is a demanding task. Existing released depth estimation networks $\mathcal{D}$ often struggle to provide ideal zero-shot predictions under various dark environments. Second, a significant domain gap exists between the depth scores and image content. Consequently, the direct fusion of depth and image features proves to be a challenging endeavor.

To address the first challenge, we introduce an encoder-decoder network denoted as $\mathcal{F}$, which predicts depth information from low-light images/frames $I_d$. We use the depth prediction from the corresponding normal-light images $I_n$ via $\mathcal{D}$ as the ground truth for supervision. This approach allows us to distill the depth prediction capability of $\mathcal{D}$ to achieve the desired depth estimation in low-light conditions (\textit{the features of pre-trained network $\mathcal{D}$ will not be employed during training and inference}). To overcome the second challenge, instead of directly fusing depth and image data, we propose a feature fusion method that involves the encoder of $\mathcal{G}$ and the depth-aware feature encoder of $\mathcal{F}$ (with the extracted features denoted as $f_g$ and $f_d$, respectively). These features are extracted from $I_d$, making them more compatible and suitable for fusion. Furthermore, these depth-aware feature priors $f_d$ not only convey the geometrical information of the target scene but also encapsulate complementary image characteristics that can enhance the features $f_g$ extracted by $\mathcal{G}$.

Once the depth-aware priors have been obtained and the fusion locations identified, the critical challenge is how to effectively carry out the fusion process. To address this, we introduce the Hierarchical Depth-Guided Feature Fusion Module (HDGFFM) at various encoder stages of $\mathcal{G}$. The depth-aware features represent the information of texture, edge, depth, and distance. 
HDGFFM facilitates feature fusion between $f_g$ and $f_d$ through a cross-attention strategy, where the depth-aware features $f_d$ serve as the query vector, while the extracted feature $f_g$ functions as the key and value vectors in the transformer's attention computation.
Compared to self-attention using $f_g$ alone, this cross-attention strategy allows us to obtain depth-aware features that encompass information beyond the image content and clearly distinguish the boundaries of different objects. These depth-aware features can then be harnessed to refine $f_g$ for enhancement, guided by the objectives of LLE. The proposed HDGFFM, with its cross-attention strategy, is adaptable to various LLE networks, making it a versatile and effective fusion method.

Extensive experiments are conducted on public LLIE and LLVE datasets with various target networks. Experimental results demonstrate that our GG-LLERF can be combined with existing LLIE and LLVE frameworks to improve their performance on varying datasets. The improvement is more significant compared to other methods using different priors, such as SKF~\cite{wu2023skf}.
In summary, our contribution is three-fold.
\begin{itemize}
    \item To the best of our knowledge, we are the first to propose using specially designed depth-aware features from input low-light images as priors for LLIE and LLVE tasks. While some previous methods rely on depth features extracted from the depth map, we uniquely assert the rationality of depth-aware features. The formulation strategy of depth-aware features is one of our key contributions.
    \item We have designed the overall structure of the Geometry-Guided Low-Light Enhancement Refine Framework, including the points at which feature fusion is conducted and the type of depth priors to be fused. This strategy is plug-and-play, allowing it to be seamlessly integrated into any low-light enhancement network.
    \item We propose the HDGFFM to perform feature fusion between depth-aware and image content features using a novel attention computation method. Compared to previous strategies, our approach integrates both self-attention and cross-attention between image and depth-aware features, along with the correlation modeling.
\end{itemize}

\section{Related Work}
\label{sec: Related Work}

\begin{figure*}[t]
	\centering
	\includegraphics[width=0.95\linewidth]{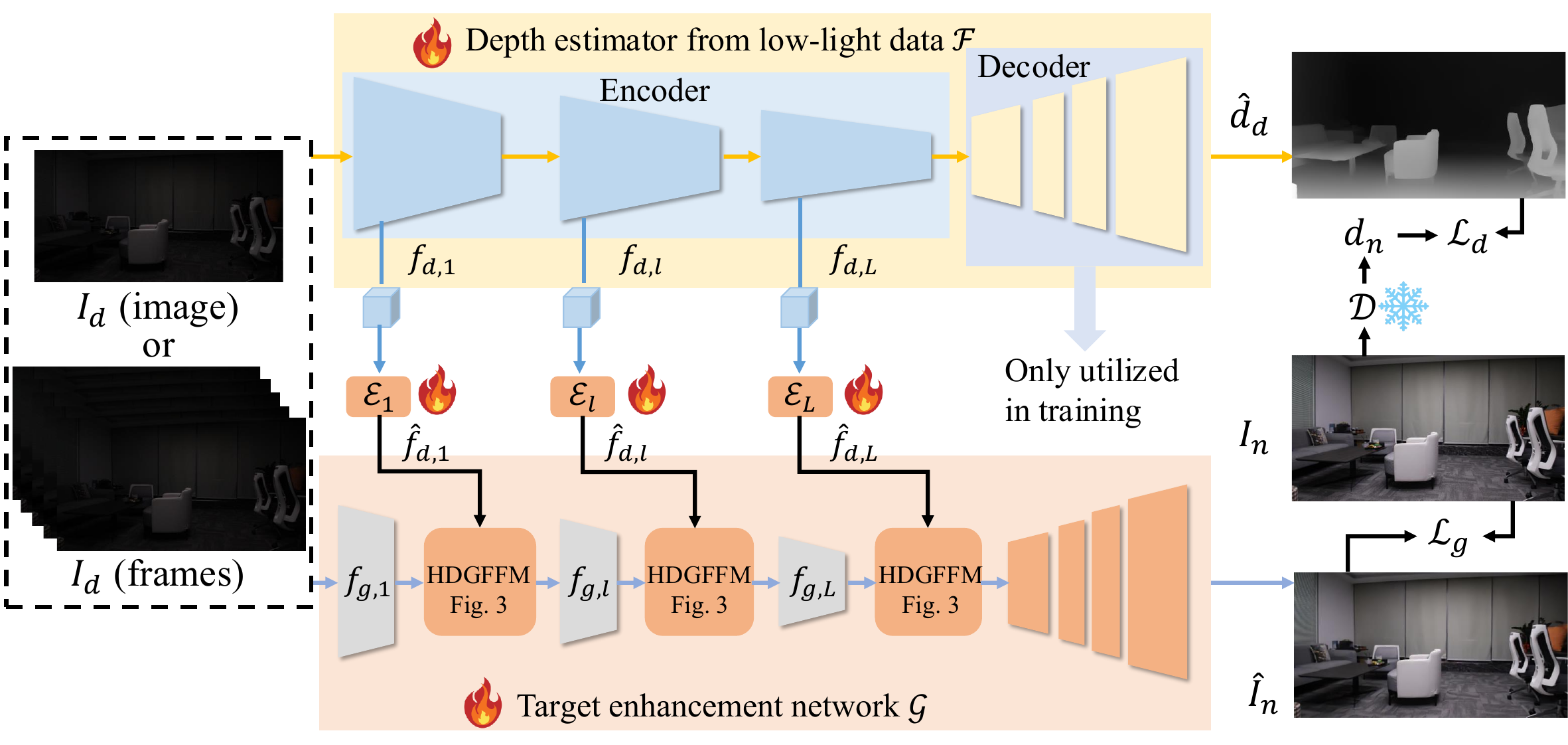}
	\vspace{-0.1in}
	\caption{
		The overall framework of GG-LLERF, which incorporates depth-aware features into the encoder of the target LLE framework (LLIE or LLVE) $\mathcal{G}$. The depth-aware features are extracted from the depth modeling network $\mathcal{F}$, which is supervised by a pre-trained depth estimator $\mathcal{D}$. $\mathcal{F}$ is trained from scratch, not pre-trained. 
		The pre-trained depth estimator $\mathcal{D}$, e.g., DPT, is just applied to the normal-light data and provides the pseudo ground truth of depth for training. $\mathcal{D}$ is frozen, and its features are not utilized. Moreover, the depth-aware features (from the encoder of $\mathcal{F}$) and image/video features are fused via our proposed Hierarchical Depth-Guided Feature Fusion Module (HDGFFM) (Sec.~\ref{sec:depth-guide}), where  Correlation-based Cross Attention is utilized (Sec.~\ref{sec:cross}).
		In contrast, the hierarchical features from the decoder of $\mathcal{F}$ are not employed during inference.
		Note that $\mathcal{G}$ can take either an image or a video as input, as it can employ both LLIE and LLVE structures. $\mathcal{F}$ is also capable of estimating depth for both images and videos, as its encoder shares the same input as that of $\mathcal{G}$, enabling effective feature fusion in HDGFFM.
	}
	\label{fig:fig1}
	\vspace{-0.2in}
\end{figure*}

\subsection{Low-light Image and Video Enhancement}
In recent years, there have been notable advancements in learning-based LLIE techniques~\cite{zamir2020learning,xu2020learning,zeng2020learning,kim2021representative,zhao2021deep,zheng2021adaptive,wang2021real,liu2021retinex,yang2021sparse,jiang2021enlightengan,yang2021band}, primarily emphasizing supervised approaches due to the availability of abundant image pairs for training. For example, MIRNet~\cite{Zamir2020MIRNet} adopts a multi-scale architecture to effectively capture and distill information at various levels. This results in improved image quality with enhanced brightness, contrast, and details, along with a reduction in noise. Fu et al.~\cite{fu2023learning} implemented illumination augmentation on a pair of images, successfully achieving self-supervised Retinex decomposition learning. This innovative approach contributes to further advancements in LLIE techniques. Zhang et al.~\cite{zhang2025} focused on using semantic segmentation priors for low-light image enhancement, enabling locally controllable enhancement. Depth information was used auxiliary to the semantic priors. Moreover, they directly obtained depth features from predicted depth maps during inference.
Wu et al.~\cite{wu2023skf} incorporated semantic information, primarily utilizing a pre-trained frozen segmentation network to obtain feature priors from low-light images, whereas the pre-trained frozen model may suffer from domain shift.
Recently, diffusion-based models also appear~\cite{wang2023exposurediffusion,yi2023diff}.

Beyond LLIE, there is a rising need for LLVE solutions. Danai et al.~\cite{triantafyllidou2020low} proposed a data synthesis mechanism that generates dynamic video pairs static datasets, formulating a LLVE framework. Wang et al.~\cite{wang2021sdsd} introduced a multi-task model capable of simultaneously estimating noise and illumination, particularly effective for videos with severe noise conditions. Xu et al.~\cite{xu2023deep} innovatively designed a parametric 3D filter tailored for enhancing and sharpening low-light videos, while cannot be utilized as a play-and-plug strategy.

While significant progress has been achieved in both LLIE and LLVE, there remains a challenge in obtaining accurate enhancement results in certain demanding scenarios. This difficulty arises from the highly ill-posed nature of directly recovering normal-light photos from these images. The ill-posed issue refers to the fact that the ground truth for an input image can vary, but only a single ground truth is provided in the training dataset. Moreover, degradation processes in real-world scenarios are often complex and unknown, leading to irreversible damage to images. As a result, various plausible outcomes can arise from the low-quality input during low-light image enhancement. Meanwhile, a representative phenomenon for the ill-posed issue occurs where some enhancement results closely approximate the target, while others do not. Adding priors helps to impose additional constraints on the enhancement process, reducing the range of plausible outcomes. Consequently, the output is closer to the distribution of the ground truth. Therefore, the incorporation of suitable priors becomes imperative to address these challenges effectively.

\subsection{Low-light Enhancement with Priors}
Given the ill-posed nature of LLE, the integration of suitable priors is essential to achieve the desired enhancement results. Recent methods have proposed to enhance the corresponding effects by incorporating multi-modal maps as unified priors. For instance, SKF~\cite{wu2023skf} utilizes semantic maps to optimize the feature space for low-light enhancement. SMG~\cite{xu2023low} adopts a generative framework that integrates edge information, enhancing the initial appearance modeling specifically designed for low-light scenarios. 

Nevertheless, the existing priors primarily operate at the 2D level, potentially falling short of fully capturing the geometric intricacies of the entire 3D scene. Recognizing the importance of incorporating 3D geometric information, which plays a pivotal role in determining illumination distribution, becomes crucial for enhancing LLE tasks. The central focus of this paper lies in exploring methods to effectively acquire and utilize 3D priors in the context of LLE.

\subsection{Differences with Current Depth-aware Restoration}
Several frameworks have been developed to incorporate depth awareness into image restoration tasks. Bao et al.~\cite{Bao2019Depth} proposed a video frame interpolation method that explicitly detects occlusions by exploiting depth information. Cheng et al.~\cite{Cheng2020ZeroShotIS} utilized depth information from a pre-trained monocular depth estimation model to restore high-resolution (HR) images in a self-supervised manner. Liet al.~\cite{Li2020Dynamic} introduced a deep neural convolutional network that leverages depth maps for dynamic scene deblurring. Lee et al.~\cite{Kim2022Geometry} proposed a geometry-aware deep video deblurring method. However, none of these methods were designed specifically for the LLE task, and their strategies for fusing depth and image information differ from our approach. They either utilize depth information as supervision or directly fuse features extracted from depth maps and images. In contrast, our paper presents a novel framework that leverages depth information by distilling depth prediction ability for low-light images (with the depth estimator trained from scratch), and fusing image content features with depth-aware image features that are not extracted from the depth map.

\section{Method}
In this section, we first provide an overview of our proposed strategy, which leverages geometrical priors (i.e., depth), in Sec.~\ref{sec:overview}. Following the overview, we delve into the structural details of our proposed Hierarchical Depth-Guided Feature Fusion (HDGFFM) in Sec.~\ref{sec:depth-guide}. This includes discussing how to obtain the desired depth prior for fusion, where the fusion takes place, and the methods employed for conducting the fusion. The following section is dedicated to describe a vital component of our fusion procedure, known as Correlation-based Cross Attention for Fusion, and can be found in Sec.~\ref{sec:cross}. Finally, we introduce the training pipeline of our strategy in Sec.~\ref{sec:loss}.

\subsection{Overview}
\label{sec:overview}
\noindent\textbf{Motivation.}
To enhance the performance of existing low-light enhancement methods, some approaches incorporate multi-modality information, such as edge and semantic maps. However, all of these incorporated priors operate at a 2D level, making them inadequate for accurately representing the corresponding real-world 3D structure. While deriving explicit 3D priors, such as point clouds and meshes, from 2D data is a highly ill-posed problem~\cite{pontes2019image2mesh,hu2021self}, there are alternative approaches to obtain pseudo-3D data in a data-driven manner, one of which is depth information.

Depth estimation has been a long-standing task, and it has recently reached new heights with the development of large models. By training on extensive datasets that encompass diverse scenes and objects, current depth estimation models (e.g., DPT~\cite{ranftl2021vision}) exhibit remarkable zero-shot performance on various images and videos. Consequently, we suggest distilling depth information from SOTA monocular depth prediction networks and integrating it into the LLE task.

\noindent\textbf{Implementation.}
In this paper, we focus on the supervised learning manner for LLE, where low-light data is denoted as $I_d$, and the normal-light data is represented as $I_n$.
As shown in Fig.~\ref{fig:fig1}, we set a lightweight depth estimation branch $\mathcal{F}$ for a given target LLE framework $\mathcal{G}$ to be improved. 
$\mathcal{F}$ takes the input of $I_d$ and output $\hat{d}_d$, as
\begin{equation}
\small
    \hat{d}_d=\mathcal{F}(I_d;\theta_f),
    \label{eq:depth_pred_low}
\end{equation}
where $\theta_f$ is the parameters to learn.
The ground truth to train $\mathcal{F}$ is obtained as the output of a pre-trained open-world depth estimator $\mathcal{D}$ (the parameter of $\mathcal{D}$ is frozen in the training stage), as $d_n=\mathcal{D}(I_n)$.

We combine the extracted features from the encoder of $\mathcal{G}$ and $\mathcal{F}$ using a cross-attention module $\mathcal{C}$. This process refines the features in the encoder of $\mathcal{G}$ by incorporating depth information from $\mathcal{F}$, ultimately enhancing the final results, denoted as $\hat{I}_n$, as
\begin{equation}
\small
    \hat{I}_n=\mathcal{G}(I_d, \mathcal{F}_e(I_d);\theta_g),
    \label{eq:encoder}
\end{equation}
where $\mathcal{F}_e(I_d)$ denotes the extracted feature with $\mathcal{F}$. The objective can be represented as 
\begin{equation}
\small
   \{ \hat{\theta}_g, \hat{\theta}_f\}=argmin(\mathcal{L}(\hat{I}_n, I_n)+\mathcal{L}(\hat{d}_d, d_n)).
    \label{eq:objective}
\end{equation}

In the following sections, we will provide the details of feature fusion.

\subsection{Hierarchical Depth-Guided Feature Fusion}
\label{sec:depth-guide}

\noindent\textbf{How to fuse?}
Assuming we can obtain the desired depth prediction $\hat{d}_d$ from the low-light input data, it remains challenging to directly integrate $\hat{d}_d$ into the enhancement process of $\mathcal{G}$ due to the fundamental difference between depth and image domains. Therefore, we propose conducting the fusion in the deep feature space, specifically between the extracted features from $\mathcal{G}$ and $\mathcal{F}$, which we denote as $f_g$ and $f_d$, respectively. This approach effectively mitigates the challenges posed by the disparity between the two data sources during fusion, as both $f_g$ and $f_d$ are derived from the same input image $I_d$.

Our experiments have revealed that conducting the fusion is more appropriate in the encoder part, as opposed to the decoder part, as done in SKF~\cite{wu2023skf}. This choice is driven by the fact that in the decoder part, the feature discrepancy tends to increase because these features are closer to the target outputs, which belong to different domains.
To elaborate, the features in the decoder of $\mathcal{G}$ represent the 2D image content, while the features in $\mathcal{F}$ capture the 3D geometrical information. Hence, it is more suitable to perform the fusion in the encoder part of both $\mathcal{G}$ and $\mathcal{F}$.

Based on the analysis provided above, we have identified both the location and the specific features to be fused within HDGFFM. In the following sections, we will elaborate on the fusion method itself.

\noindent\textbf{The fusion pipeline in HDGFFM.}
In a typical low-light enhancement network, a hierarchical encoder structure is employed. To ensure compatibility between the feature sets $f_g$ and $f_d$, the encoder of $\mathcal{F}$ is also designed in a pyramidal manner. Consequently, both $f_g$ and $f_d$ consist of a sequence of features, which we denote as $f_g=\{f_{g,l}\}, l\in[1,L]$, and $f_d=\{f_{d,l}\}, l\in[1,L]$, where $L$ represents the number of layers in the encoder.
To further standardize the channel representation of $f_g$ and $f_d$, we incorporate a depth-aware embedding module, denoted as $\mathcal{E}_l$, in each layer. These modules are responsible for processing the depth-aware features within the respective layers, as
\begin{equation}
\small
    \hat{f}_{d,l}=\mathcal{E}_l(f_{d,l}).
    \label{eq:embedding}
\end{equation}
The depth-aware embedding modules adjust channels without information compression. Therefore, they are designed to be lightweight, consisting of a single layer of linear convolution.

The fusion process is carried out through our cross-attention mechanism, which will be elaborated on in the next section.

\begin{figure}[t]
    \centering
    \includegraphics[width=0.8\linewidth]{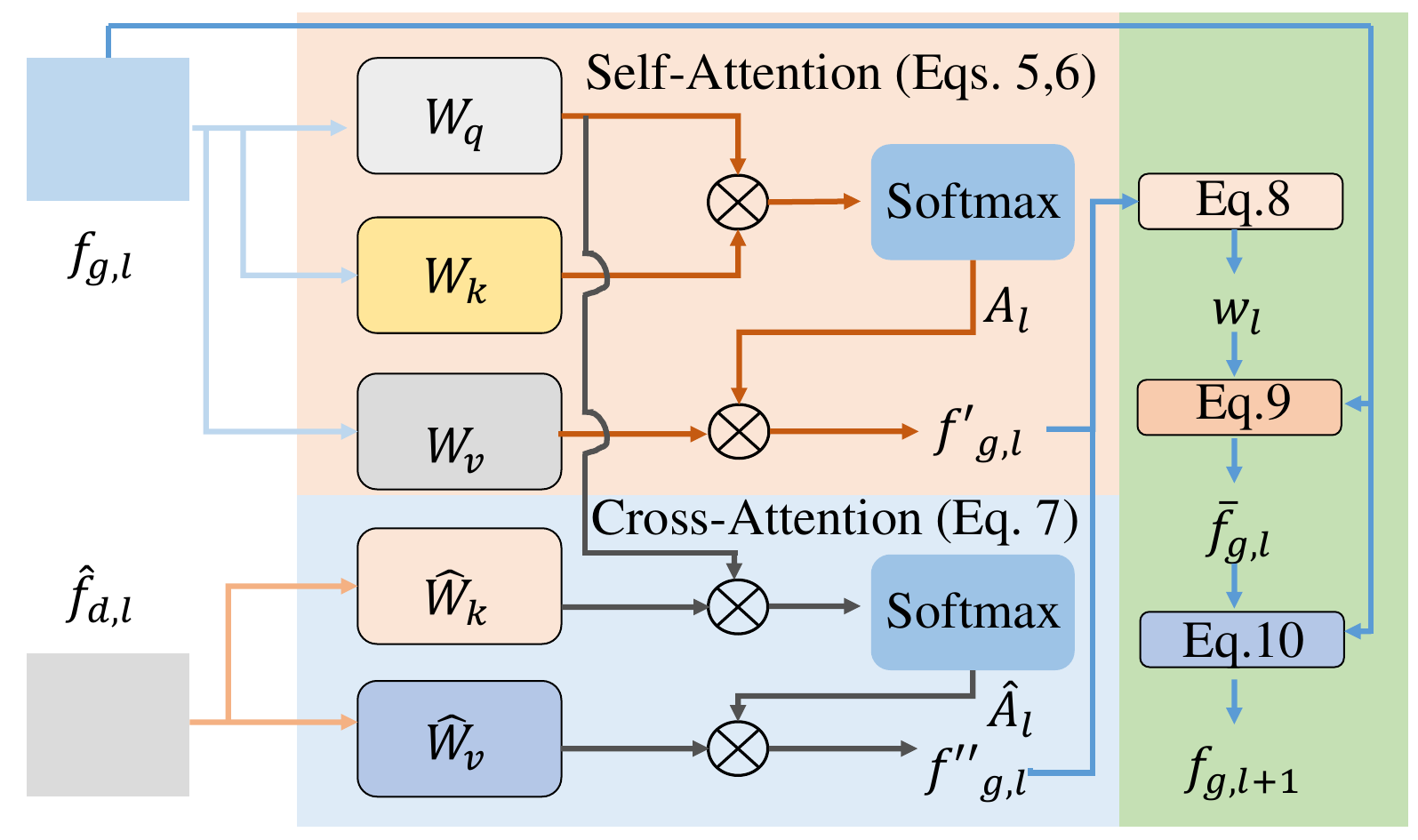}
    \vspace{-0.1in}
    \caption{The overview of ``Correlation-based Cross Attention for Fusion''. The attention map is computed in both self-attention (Eq~\ref{eq:self}) and cross-attention between different domains (Eq~\ref{eq:cross-attention}).}
    \label{fig:fig2}
    \vspace{-0.2in}
\end{figure}

\subsection{Correlation-based Cross Attention for Fusion}
\label{sec:cross}
\noindent\textbf{Feature fusion strategy.}
Several strategies have been employed to fuse cross-domain features for low-light enhancement. In SMG~\cite{xu2023low}, synthesized edge information serves as a condition for generating feature normalization and convolution parameters. However, this operation can be computationally expensive, especially when dealing with high-resolution images. On the other hand, SKF~\cite{wu2023skf} computes the similarity between image and semantic features and utilizes this similarity to modulate the original image features, i.e., they use image, semantic, and image features as query, key, and value features in the fusion. It's worth noting that this strategy may not effectively incorporate new information, as the semantic features are primarily used to facilitate the computation of an attention matrix rather than being integrated with the image features themselves. And our method's superiority to these methods are shown in Table. \ref{table:quantitative1}.

In this paper, as shown in Fig.~\ref{fig:fig2}, our approach involves directly guiding the depth-aware features into the pathways of $\mathcal{G}$. Additionally, we integrate the results of the cross-attention computation with the outputs of the self-attention computation, based on a feature correlation measurement.

\noindent\textbf{Fusion computation details.}
Suppose our goal is to fuse features in the $l$-th layer, denoted as $f_{g,l}$ and $\hat{f}_{d,l}$. Both of these feature maps have the same shape of $H\times W\times C$, where $H$ represents the feature height, $W$ the feature width, and $C$ the feature channel. To perform the feature computation, projection matrices for query, key, and value~\cite{vaswani2017attention} are established for both $f_{g,l}$ and $\hat{f}_{d,l}$. For $f_{g,l}$, these projection matrices are labeled as $W_q$, $W_k$, and $W_v$, while for $\hat{f}_{d,l}$, only the key and query matrices are required, designated as $\hat{W}_k$ and $\hat{W}_v$, respectively.

The self-attention is first conducted for $f_{g,l}$, by computing the self-similarity matrix, as
\begin{equation}
\small
    A_{l}=Softmax(W_q(f_{g,l}) \times W_k(f_{g,l})^T \times \tau),
    \label{eq:self}
\end{equation}
where $W_q(f_{g,l})$ and $W_k(f_{g,l})$ has the shape of $C \times (H\times W)$, $T$ is the transpose operation, $\tau$ is a temperature hyper-parameter, $\times$ is the matrix multiplication operation, and $A_{l}$ has the shape of $C\times C$. We utilize a channel-wise attention map to maintain low computational costs, though an attention map with spatial dimensions is also compatible with our framework.
The output of the self-attention is obtained as
\begin{equation}
\small
    f'_{g,l}=A_{l} \times W_v(f_{g,l}),
    \label{eq:self2}
\end{equation}
where $f'_{g,l}$ has the shape of $C \times (H\times W)$.

Regarding cross-attention, the similarity matrix can be computed between the image and depth-aware features, and the final output can be determined based on this computed similarity matrix.
The procedure can be written as
\begin{equation}
\begin{aligned}
\small
    &\hat{A}_{l}=Softmax(W_q(f_{g,l}) \times \hat{W}_k(\hat{f}_{d,l})^T \times \tau),\\
    &f''_{g,l}=\hat{A}_{l} \times \hat{W}_v(\hat{f}_{d,l}),
\end{aligned}
\label{eq:cross-attention}
\end{equation}
where $\hat{A}_{l}$ and $f''_{g,l}$ have the same shape as that of $A_{l}$ and $f'_{g,l}$.

Note that we adopt channel-wise attention due to its lower computational complexity compared to spatial-wise attention. We give a comprehensive complexity analysis here.
The attention computation process consists of four main components: feature projection, attention map computation between the query and key features, the softmax operation, and the multiplication between the attention map and the value feature, as outlined in Eq~\ref{eq:self} and Eq~\ref{eq:self2} of the main paper. Notably, the computational complexity remains the same for both self-attention (Eq~\ref{eq:self}/\ref{eq:self2}) and cross-attention (Eq~\ref{eq:cross-attention}). We assume the input feature map has a shape of $H \times W \times C$.

\begin{itemize}
    \item \textbf{Feature projection in Eq~\ref{eq:self}}: Our projection layer includes $W_q$, $W_k$ (or $\hat{W}_k$), and $W_v$ (or $\hat{W}_v$). These are obtained via convolutional layers: the first layer is a $1 \times 1$ convolution that transforms the channel dimension from $C$ to $3C$, and the second is a depthwise convolution with a kernel size of $3 \times 3$ and group size $3C$, maintaining the channel dimension at $3C$. The computational cost of the first layer is $H\times W \times C \times 3C$, and that of the second layer is $H\times W \times 3 \times 3 \times 3C$. Therefore, the total computational complexity is $\mathcal{O}((27C+3C^2)HW)$.

    \item \textbf{Attention map computation in Eq~\ref{eq:self}}: 
    For channel-wise attention, the attention map has a shape of $C \times C$, obtained by multiplying two features of sizes $C\times (HW)$ and $(HW)\times C$. The resulting complexity is $\mathcal{O}(C^2HW)$. In contrast, spatial-wise attention produces an attention map of size $(HW) \times (HW)$, formed by multiplying features of sizes $(HW)\times C$ and $C\times (HW)$, leading to a computational complexity of $\mathcal{O}(CH^2W^2)$.

    \item \textbf{$\times \tau$ in Eq~\ref{eq:self}}: The attention map is scaled by a temperature factor $\tau$, which involves an element-wise multiplication. This operation introduces an additional computational complexity of $\mathcal{O}(C^2)$ for channel-wise attention and $\mathcal{O}(H^2W^2)$ for spatial-wise attention, respectively.

    \item \textbf{Softmax operation in Eq~\ref{eq:self}}: The softmax operation is applied row-wise to the attention map and involves three main steps: computing the exponent of each element, summing these exponents across the row, and dividing each element by the sum. Consequently, the computational complexity is $\mathcal{O}(3C^2)$ for channel-wise attention and $\mathcal{O}(3H^2W^2)$ for spatial-wise attention.

    \item \textbf{Multiplication between the attention map and value part in Eq~\ref{eq:self2}}: After obtaining the attention map, it is multiplied with the value feature to produce the output. For channel-wise attention, this step has a computational complexity of $\mathcal{O}(C^2HW)$; for spatial-wise attention, the complexity is $\mathcal{O}(CH^2W^2)$.

\end{itemize}
In summary, the computational complexity of our channel-wise attention mechanism is $\mathcal{O}\big((27C+3C^2)HW + C^2HW + C^2+3C^2+C^2HW\big) = \mathcal{O}\big((27C+5C^2)HW + 4C^2\big)$, where the dominant term is $\mathcal{O}((27C+5C^2)HW)$ under the assumption that $C \ll H$ and $C \ll W$.
In contrast, the complexity of the spatial-wise attention mechanism is
$\mathcal{O}\big((27C+3C^2)HW + CH^2W^2+H^2W^2+3H^2W^2+CH^2W^2\big) = \mathcal{O}\big((27C + 3C^2)HW + (4 + 2C)H^2W^2\big)$, which is significantly higher than that of the channel-wise attention.

As both $f_{g,l}$ and $\hat{f}_{d,l}$ are derived from the encoder process of $I_d$ and exhibit homogeneity, as discussed in Sec.~\ref{sec:depth-guide}, $f'_{g,l}$ and $f''_{g,l}$ also exhibit small heterogeneity between them. To further minimize the disparity between $f'_{g,l}$ and $f''_{g,l}$, we introduce a method to model the correlation between these two sets of features via a fusion network, and then utilize this correlation as a weighting factor for the final fusion process.
The correlation is obtained as 
\begin{equation}
\small
    w_l=Sigmoid(\mathcal{O}(f'_{g,l} \oplus f''_{g,l})),
    \label{eq:sigmoid}
\end{equation}
where $\oplus$ is the channel concatenation operation and $\mathcal{O}$ is fusion network.
The fusion output can be written as
\begin{equation}
\small
    \bar{f}_{g,l}=w_l \cdot f''_{g,l} + f'_{g,l}+f_{g,l},
    \label{eq:fusion-eq}
\end{equation}
where $\cdot$ is the element-wise multiplication.
Moreover, the final outputs of HDGFFM is obtained with the attention outputs and the feed-forward network, as
\begin{equation}
\small
    F_{g,l}=f_{g,l+1}=\mathcal{H} (\bar{f}_{g,l})+f_{g,l},
    \label{eq:fusion-eq2}
\end{equation}
where $\mathcal{H}$ denotes the feed-forward network.

\subsection{Training Strategy}
\label{sec:loss}
In Eq~\ref{eq:objective}, two objectives are considered: the restoration loss for enhancement and the depth supervision loss to obtain accurate depth priors. The restoration loss, denoted as $\mathcal{L}_{g}$, can employ the same loss terms as those used for the target model $\mathcal{G}$.
On the other hand, the depth supervision loss is implemented as the $L_2$ distance between the predicted depth values and the ground truth depth information, as
\begin{equation}
\small
    \mathcal{L}_d=\Vert \hat{d}_d - d_n \Vert_2.
\label{eq:depth loss}
\end{equation}
The final loss to train our strategy can be written as
\begin{equation}
\small
    \mathcal{L}=\mathcal{L}_{g}+\lambda \mathcal{L}_{d},
    \label{eq:final}
\end{equation}
where $\lambda$ is the loss weight, which is robust across different target scenarios. 
The training procedure of our method is summarized in Algorithm~\ref{alg:ours}.

\begin{algorithm}[t]
\caption{The training procedure of our method.}
\label{alg:ours}
\begin{algorithmic}
\renewcommand{\algorithmicrequire}{\textbf{Input:}}
\renewcommand{\algorithmicensure}{\textbf{Output:}}

\algnewcommand\algorithmicforeach{\textbf{for each}}
\algdef{S}[FOR]{ForEach}[1]{\algorithmicforeach\ #1\ \algorithmicdo}
    \Require{Training dataset with pairs $\{(I_d, I_n)\}$, initialized target enhancement network $\mathcal{G}$ with parameters $\theta_g$ and depth estimator for low-light data $\mathcal{F}$ with parameters $\theta_f$, pre-trained depth estimation network $\mathcal{D}$, current training iteration $T$, maximum training iteration $T_{max}$},

\While {$T < T_{max}$}
\State Sample training batch $(I_d, I_n)$ from the dataset
\State Use the frozen $\mathcal{D}$ to predict $d_n=\mathcal{D}(I_n)$
\State Conduct the depth prediction from $I_d$ to obtain $\hat{d}_d$ as shown in Eq~\ref{eq:depth_pred_low}, and obtain the feature of each encoder layer as $f_d=\{ f_{d,l} \}, l\in [1, L]$
\State Extract the image features from $I_d$ with the encoder of $\mathcal{G}$, as $f_g=\{ f_{g,l} \}, l\in [1, L]$
\State Forward $\{ f_{d,l} \}, l\in [1, L]$ to the depth-aware embedding module $\mathcal{E}_l$, to obtain $\{ \hat{f}_{d,l} \}, l\in [1, L]$, as shown in Eq~\ref{eq:embedding}
\State Obtain the fusion results of $\{ f_{g,l} \}, l\in [1, L]$ and $\{ \hat{f}_{d,l} \}, l\in [1, L]$ via Eqs~\ref{eq:self}, \ref{eq:self2}, \ref{eq:cross-attention}, \ref{eq:sigmoid}, \ref{eq:fusion-eq}, and \ref{eq:fusion-eq2}, as $\{ F_{g,l} \}, l\in [1, L]$
\State Compute the output images with the fusion results and the decoder of $\mathcal{G}$, as $\hat{I}_n$
\State Compute the restoration loss as $\mathcal{L}_g$ with $\hat{I}_n$
\State Compute the depth supervision loss as $\mathcal{L}_d$ with $\hat{d}_d$ and $d_n$, as shown in Eq~\ref{eq:depth loss}
\State Get the final loss as shown in Eq~\ref{eq:final}, and update the network of $\mathcal{G}$ and $\mathcal{F}$ simultaneously
\State $T=T+1$
\EndWhile
\Ensure{trained model $\mathcal{G}$ and $\mathcal{F}$ }
\end{algorithmic}
\end{algorithm}

\noindent\textbf{The applicability of our strategy for LLVE.}
While our method is evidently well-suited for LLIE, we emphasize that it is also applicable to the LLVE task, although it is not specifically designed for it. 
However, it is rational for LLVE because the pre-trained depth estimator $\mathcal{D}$ we use is typically trained on video data, enabling it to produce stable depth estimates without noticeable flickering in normal videos. For example, see the consistent depth results demonstrated in the DepthAnything project (https://depth-anything.github.io/ and https://depth-anything-v2.github.io/).
Therefore, using such models to provide frame-wise depth supervision does not cause significant temporal inconsistency. This justifies our application of the method to LLVE videos, resulting in the expected improvement.

\section{Experiment}
\subsection{Experimental Settings}
\label{sec:ExperimentalSetting}

\noindent\textbf{Datasets.} 
We assessed the proposed framework using various LLIE/LLVE datasets, which include SDSD-indoor~\cite{wang2021sdsd}, SDSD-outdoor~\cite{wang2021sdsd}, LOLv2~\cite{yang2021sparse}, SMID~\cite{chen2019seeing}, SID~\cite{chen2018learning}, and DAVIS2017~\cite{pont20172017}.
\begin{itemize}
    \item \textbf{SDSD}: The videos in SDSD were captured in dynamic pairs using electromechanical equipment. 
    \item \textbf{LOLv2}: LOLv2 is subdivided into LOLv2-real and LOLv2-synthetic, and they are commonly utilized in LLIE.
    \item \textbf{SMID and SID}: SMID is a static video dataset consisting of frames captured with short exposures and ground truth obtained using long exposures. The collection approach for SID is similar to that of SMID but introduces additional challenges with extreme situations. For both SMID and SID, we use full images and convert RAW data to RGB since our work focuses on LLE in the RGB domain. 
    \item \textbf{DAVIS2017}: The utilization of the DAVIS dataset for LLE was initially proposed by \cite{zhang2021learning}. It synthesizes low-light and normal-light video pairs with dynamic motions. In comparison to \cite{zhang2021learning}, we further incorporate the degradation of noise into low-light videos, in addition to invisibility, aligning more closely with real-world low-light videos.
\end{itemize}
All training/testing splits adhere to the guidelines specified in the original papers.

\begin{table*}[tb]
\begin{center}
\large
\caption{Quantitative comparison in the LLVE task on SDSD and SMID. Note that the batch size of SDSD in the SMID dataset is 12 that is smaller than the original paper due to limited computation resources. ``T.C." means the temporal consistency evaluation using \cite{liu2023evalcrafter} }
\label{comparison1}
		\resizebox{\linewidth}{!}{
        \begin{tabular}{l|ccccc|ccccc|ccccc}
	\toprule
	\multirow{2}{*}{Methods}& \multicolumn{5}{c|}{SDSD-indoor} &\multicolumn{5}{c|}{SDSD-outdoor}&\multicolumn{5}{c}{SMID} \\
	\cline{2-16}
	 & PSNR & SSIM &LPIPS &RMSE &T.C. & PSNR & SSIM &LPIPS &RMSE &T.C. & PSNR & SSIM &LPIPS &RMSE&T.C.  \\
	\hline
	DP3DF~\cite{xu2023deep}&28.90 &0.880 &0.235 &9.290&4.16e-03 &27.24 & 0.794 &0.336 &12.303 &2.26e-03 &25.60 &0.727 &0.339 &14.225 &6.21e-04 \\
        Ours &\textbf{30.95} &\textbf{0.897} &\textbf{0.231} &\textbf{7.280}&\textbf{4.48e-03}&\textbf{27.89}&\textbf{0.804}&0.343&\textbf{10.708}&\textbf{2.28e-03}&\textbf{26.35}&\textbf{0.743}&0.345 &\textbf{13.677}&\textbf{4.56e-04}\\
        \hline
	SDSD~\cite{wang2021sdsd} &25.81&0.749&0.328 &13.468&4.53e-03&25.17&0.763&0.244 &14.421&2.86e-03 &24.13 &0.715&0.294 &16.870&4.86e-03 \\
        Ours &\textbf{27.53} &\textbf{0.766} &\textbf{0.311} &\textbf{10.970}&\textbf{4.31e-03}&\textbf{27.26}&\textbf{0.794}&\textbf{0.210} &\textbf{11.352}&\textbf{2.06e-03}&\textbf{26.04}&\textbf{0.746}&\textbf{0.238} &\textbf{13.788}&\textbf{1.78e-03}\\
	\bottomrule[1pt]
        \end{tabular}}
\end{center}
\end{table*}     

\begin{table}[tb]
\caption{Quantitative comparison in the LLVE task on the DAVIS2017 dataset.}
\label{comparison1-2}
\centering
\resizebox{\linewidth}{!}{
\begin{tabular}{l|p{1cm}<{\centering}|p{1cm}<{\centering}|p{1cm}<{\centering}|p{1cm}<{\centering}}
	\toprule
    \centering
	\multirow{2}{*}{Methods}&\multicolumn{4}{c}{DAVIS2017} \\
	\cline{2-5}
	 & PSNR & SSIM &LPIPS &RMSE  \\
	\hline
	StableLLVE~\cite{zhang2021learning} &26.39&0.977&0.283&61.589\\	
	Ours &\textbf{28.04 }&\textbf{0.981 }&0.284&\textbf{59.011}\\
	\bottomrule[1pt]
        \end{tabular}}
\end{table}

\begin{table}
\centering
\caption{The experimental results on different categories of data, involving DP3DF on SDSD-outdoor.}
\label{classified test}
\resizebox{1.0\linewidth}{!}{
\begin{tabular}{l|cc|cc}
	\toprule
	\multirow{1}{*}{Methods}& \multicolumn{1}{c}{PSNR}& \multicolumn{1}{c}{SSIM}& \multicolumn{1}{c}{LPIPS}& \multicolumn{1}{c}{RMSE}\\
	\cline{2-5}
	\hline
	Building in Baseline
	&24.68&0.743&0.011&14.88\\
        Building in ours
        &\textbf{24.76}&\textbf{0.751}&\textbf{0.009}&\textbf{14.76}\\
    \hline
	car in Baseline
	&29.65&0.843&0.004&8.74\\
        car in ours
&\textbf{29.83}&\textbf{0.846}&\textbf{0.003}&\textbf{8.57}\\
    \hline
        floor in Baseline
	&24.58&0.744&0.010&15.06\\
        floor in ours
	&\textbf{24.94}&\textbf{0.757}&\textbf{0.009}&\textbf{14.44}\\
    \hline
        plant in Baseline
	&21.78&0.622&0.002&20.96\\
        plant in ours
	&\textbf{25.58}&\textbf{0.718}&\textbf{0.007}&\textbf{13.47}\\
    \bottomrule 
	Average in Baseline &25.17 &0.738&0.007&14.91  \\
        Average in ours &\textbf{26.28} &\textbf{0.768} &\textbf{0.007}&\textbf{12.81} \\
        \bottomrule 
\end{tabular}}
\end{table}

\begin{table}[t]
\centering
\caption{The evaluation of temporal consistency using subjective scores. The best temporal consistency is marked as 5, while the worst is marked as 1.}
\label{subjective index}
\resizebox{1.0\linewidth}{!}{
\begin{tabular}{l|p{2.5cm}<{\centering}|p{2.5cm}<{\centering}}
	\toprule
        Methods& {SDSD-indoor} &{SDSD-outdoor}\\
	\cline{2-3}
        \hline
	DP3DF
	&3.48&2.23\\
        Ours
        &\textbf{3.67}&\textbf{2.34}\\
        \hline
        SDSD
	&3.92&2.85\\
        Ours
        &\textbf{4.11}&\textbf{3.49}\\
        \bottomrule
 \end{tabular}}
\end{table}

\begin{table}[t]
\centering
\caption{Model parameter number (M), FLOPs (T), and running time (milliseconds) in LLVE/LLIE task for DP3DF, SDSD, StableLLVE, Restormer, SNR, Retinexformer, HVI-CIDNET and IAT.
The running time and FLOPs are computed with a size of 512$\times$960.}
\label{params and runtime}
\resizebox{1.0\linewidth}{!}{
\begin{tabular}{l|p{1.5cm}<{\centering}|p{1.5cm}<{\centering}|p{1.5cm}<{\centering}}
	\toprule
Methods& Parameter & FLOPs & Time\\
	\cline{2-3}
        \hline
	DP3DF
	&28.86&4.65&13.54\\
        ours
        &\textbf{33.32}&\textbf{4.94}&\textbf{21.44}\\
        \hline
        SDSD
	&4.30&0.34&69.70\\
        ours
        &\textbf{6.67}&\textbf{1.32}&483.00\\
        \hline
        StableLLVE
	&4.32&0.32&48.20\\
        ours
        &\textbf{8.60}&\textbf{0.70}&209.50\\
        \hline
	Restormer
	&26.13&1.06&261.20\\
        ours
        &\textbf{31.41}&\textbf{1.20}&\textbf{309.64}\\
        \hline
        SNR
	&39.12&0.18&37.50\\
        ours
        &\textbf{41.49}&\textbf{0.38}&108.10\\
        \hline
        Retinexformer
	&1.61&0.13&85.80\\
        ours
        &15.65&\textbf{0.93}&\textbf{184.90}\\
        \hline
        HVI-CIDNET
	&1.97&0.01&117.40\\
        ours
        &\textbf{2.75}&\textbf{0.02}&\textbf{134.10}\\
        \hline
        IAT
	&0.09&0.01&33.70\\
        ours
        &\textbf{0.16}&\textbf{0.07}&\textbf{87.40}\\
        \bottomrule
 \end{tabular}}
\end{table}

\vspace{1mm}
\noindent\textbf{Metrics.} 
To assess the performance of various frameworks, we employ full-reference image quality evaluation metrics. Specifically, we utilize peak signal-to-noise ratio (PSNR) and structural similarity (SSIM)~\cite{wang2004image}.

\vspace{1mm}
\noindent\textbf{Compared methods.} 
To validate the effectiveness of our method, we conduct comparisons with a set of SOTA methods for both LLIE and LLVE tasks. For video tasks, we evaluate against DP3DF~\cite{xu2023deep}, SDSD~\cite{wang2021sdsd}, and StableLLVE~\cite{zhang2021learning}, following their official settings and objectives except for batch size. For LLIE tasks, we include two representative frameworks: 
Restormer~\cite{zamir2022restormer} and SNR~\cite{xu2022snr}.
Compared to the baseline, the modification brought by applying our method includes the extraction of depth-aware image features and fusion layers during inference, as well as the additional training of the depth estimator during training.
Moreover, KinD~\cite{zhang2019kindling}, URetinex-Net++~\cite{wu2022uretinex}, LLFlow-L-SKF~\cite{wu2023skf}, Retinexmamba~\cite{bai2024retinexmamba}, Wakeup-Darkness~\cite{zhang2025}, Zero-DCE~\cite{Zero-DCE}, RUAS~\cite{liu2021ruas}, SCI~\cite{ma2022toward}, NeRCO~\cite{yang2023implicit}, and Clip-LIE~\cite{liang2023iterative} are included to show the SOTA performance that can be achieved by our method.

\vspace{1mm}
\noindent\textbf{Implementation details.} 
We implement our framework in PyTorch and conducted experiments on an NVIDIA A100 GPU and an NVIDIA A40 GPU. Our implementation is based on the released code of the baseline networks, and we ensure the use of the same training settings for both the baseline and our method except for batch size.

\begin{table*}[tb]
    \caption{Quantitative comparison in LLIE task on SDSD-indoor, SDSD-outdoor, SMID, LOLv2-real, LOLv2-synthetic, and SID. Our mothod outperforms almost all baselines consistently. The dataset to train ``SNR" is the original dynamic SDSD dataset, which is the same as other methods in this paper and different from~\cite{xu2022snr} using its static version. Note that the batch size of Restormer is smaller than the original paper due to limited computation resources. }
    \label{table:quantitative1}
    \renewcommand\arraystretch{1.15}
    \centering
    \resizebox{\textwidth}{!}{
    \begin{tabular}{l|cc|cc|cc|cc|cc|cc}
		\toprule
	\multirow{2}{*}{Methods}&\multicolumn{2}{c|}{SDSD-indoor} &
			\multicolumn{2}{c|}{SDSD-outdoor} & 
			\multicolumn{2}{c|}{SMID} & 
			\multicolumn{2}{c|}{LOLv2-real} & 
			\multicolumn{2}{c|}{LOLv2-synthetic}& 
			\multicolumn{2}{c}{SID}\\
        \cline{2-13}
	& PSNR & SSIM& PSNR & SSIM & PSNR & SSIM& PSNR & SSIM& PSNR & SSIM& PSNR & SSIM\\
        \cline{1-13}
		Retinexformer~\cite{cai2023retinexformer} &25.00 &0.894 &25.69 &0.780 &28.66 &0.805 &19.60 &0.770 &24.20 &0.926 &21.59 &0.560 \\
            Ours &\textbf{25.12} &\textbf{0.903} &\textbf{27.53}&\textbf{0.801}  & \textbf{28.72}&\textbf{0.810} &\textbf{20.13}&\textbf{0.770}&\textbf{24.97}&\textbf{0.928}&\textbf{22.17}&\textbf{0.566} \\
        \hline
			Restormer~\cite{zamir2022restormer} &27.11 &0.923 &29.53 &0.838 &28.68 &0.810 &19.49 &0.852 &25.74 &0.943 &22.99 &0.533\\
            Ours &\textbf{27.40} &\textbf{0.924} &\textbf{29.91}&\textbf{0.850}  & \textbf{29.28}&\textbf{0.818} &\textbf{19.95}&\textbf{0.853}&\textbf{26.13}&\textbf{0.947}&\textbf{23.77}&\textbf{0.634}\\
        \hline
			SNR~\cite{xu2022snr} &26.15 &0.914 &28.81 &0.848 &28.21 &0.801 &19.98 &0.828 &23.93 &0.929 &22.69 &0.616 \\
            SNR+SKF~\cite{wu2023skf}&23.32 &0.876 &27.16 &0.812 &27.68 &0.793 &21.65 &0.809 &17.70 &0.785 &20.32 &0.563 \\
            Ours &\textbf{26.86} &\textbf{0.924} &\textbf{29.38}&\textbf{0.858}  & \textbf{29.01}&\textbf{0.813} &20.88&\textbf{0.849} & \textbf{25.53}&\textbf{0.940} &\textbf{26.53} &\textbf{0.678} \\
			\bottomrule 
\end{tabular}}
\end{table*}

\begin{table*}[tb]
    \caption{The quantitative comparisons with the representative low-light image enhancement baseline and its version enhanced by other priors.}
    \label{table:Lpips and RMSE for SNR}
    \renewcommand\arraystretch{1.15}
    \centering
    \resizebox{\textwidth}{!}{
    \begin{tabular}{l|cc|cc|cc|cc|cc|cc}
		\toprule
	\multirow{2}{*}{Methods}&\multicolumn{2}{c|}{SDSD-indoor} &
			\multicolumn{2}{c|}{SDSD-outdoor} & 
			\multicolumn{2}{c|}{SMID} & 
			\multicolumn{2}{c|}{LOLv2-real} & 
			\multicolumn{2}{c|}{LOLv2-synthetic}& 
			\multicolumn{2}{c}{SID}\\
        \cline{2-13}
	& LPIPS & RMSE& LPIPS & RMSE & LPIPS & RMSE& LPIPS & RMSE& LPIPS & RMSE& LPIPS & RMSE\\
        \cline{1-13}
        \hline
		SNR~\cite{xu2022snr} &0.068 &14.061 &0.139 &9.961 &0.148 &10.903 &0.066 &30.265 &0.034 &19.317 &0.202 &20.471 \\
            SNR+SKF~\cite{wu2023skf}&0.084&18.039 &\textbf{0.135} &13.643 &0.160 &11.415 &0.101 &22.957 &0.104 &39.688 &0.287 &27.176 \\
            Ours &\textbf{0.063} &\textbf{13.028} &\textbf{0.135}&\textbf{9.429}  & \textbf{0.142}&\textbf{10.138} & 0.069&29.788 & \textbf{0.027}&\textbf{16.398}& \textbf{0.154}&\textbf{13.630}\\
			\bottomrule 
\end{tabular}}
\end{table*}

\begin{table*}[tb]
    \caption{The quantitative comparisons between the representative low-light image enhancement baseline and its version enhanced by our method.}
    \label{table:HVI and IAT}
    \renewcommand\arraystretch{1.15}
    \centering
    \resizebox{\textwidth}{!}{
    \begin{tabular}{l|cccc|cccc|cccc}
		\toprule
	\multirow{2}{*}{Methods}&\multicolumn{4}{c|}{SDSD-indoor} &
			\multicolumn{4}{c|}{SDSD-outdoor} & 
			\multicolumn{4}{c}{SMID} 
			\\
        \cline{2-13}
	& PSNR & SSIM& LPIPS & RMSE & PSNR & SSIM& LPIPS & RMSE& PSNR & SSIM& LPIPS & RMSE\\
        \cline{1-13}
			HVI-CIDNET~\cite{yan2025hvi} &22.71 &0.903 &0.142 &24.225 &26.07 &0.806 &0.292 &12.570&26.41 &0.775 &0.223 &13.612 \\
            Ours &\textbf{25.81} &\textbf{0.913} &\textbf{0.122}&\textbf{16.323}  & \textbf{27.02}&\textbf{0.866} &\textbf{0.215}&\textbf{11.359}&\textbf{26.64}&0.614&0.264&\textbf{11.862} \\
        \hline
			IAT~\cite{Cui_2022_BMVC} &15.28 &0.694 &0.221 &44.740 &15.01 &0.692 &0.348 &63.240 &25.07 &0.754 &0.171 &15.010 \\
            Ours &\textbf{17.58} &\textbf{0.729} &\textbf{0.190}&\textbf{39.540}  & \textbf{16.83}&\textbf{0.670} &\textbf{0.159}&\textbf{42.150}&\textbf{25.34}&\textbf{0.758}&\textbf{0.163}&\textbf{14.610} \\
	\bottomrule 
\end{tabular}}
\vspace{-0.1in}
\end{table*}

\subsection{Comparison}
\label{sec:Quantitative Evaluation}
\noindent\textbf{Quantitative comparison on LLVE tasks.}
The quantitative comparisons for LLVE baselines and their versions with our depth guidance are presented in Tables.~\ref{comparison1} and \ref{comparison1-2}. Specifically, in DP3DF, our method demonstrates improvements of 2.05dB in PSNR and 0.017 in SSIM on the SDSD-indoor dataset. Notably, within the SDSD framework, our method achieves an enhancement of 1.72dB in PSNR and 0.017 in SSIM on the SDSD-indoor dataset. Furthermore, in the StableLLVE framework evaluated on the DAVIS2017 dataset, our results exhibit a significant increase of 1.65dB in PSNR, showcasing substantial improvement, particularly in various dynamic scenes.
In addition, we conduct a classification of the dataset to evaluate the performance on different regions. Using the pretrained semantic classification model CLIP~\cite{radford2021learning} and manual refinement, we divide our evaluation dataset into different categories, including floor, building, plants, and car. As shown in Table~\ref{classified test}, our approach has better performance on all categories.

Moreover, we further employ the evaluation metric for temporal consistency, using warping error~\cite{liu2023evalcrafter}. The results are displayed in Tables.~\ref{comparison1} and \ref{comparison1-2}, showing that the temporal consistency of the results enhanced by our method is improved for most cases. Moreover, we believe it's important to include a subjective evaluation of temporal consistency. To this end, we conducted a user study, inviting more than 30 participants. The results, shown in Table.~\ref{subjective index}, reveal that participants gave higher scores for temporal consistency in our method.

Additionally, we provide metrics for ``model parameters'', ``FLOPs", and ``running time'' for both baselines and our method in Table.~\ref{params and runtime}. The results illustrate that the improvements achieved by our method do not come at the severe expense of efficiency. 

\vspace{1mm}
\noindent\textbf{Quantitative comparison on LLIE tasks.}
For the evaluation of LLIE, we conduct experiment on a variety of datasets, showing the effectiveness of our strategy more comprehensively and convincingly. The results are displayed in Table.~\ref{table:quantitative1}, where we can see that almost all baselines are improved by involving our strategy without additional heavy computation cost.
Within the Restormer framework, our method achieves enhancements of 0.012 in SSIM on the SDSD-outdoor dataset. Additionally, in the SNR framework evaluated on the LOLv2-synthetic dataset, our results demonstrate a notable increase of 1.6dB in PSNR and 0.011 in SSIM, indicating substantial improvements across different scenarios.
Moreover, we include the comparisons with more baselines and metrics (LPIPS and RMSE), as displayed in Tables.~\ref{table:Lpips and RMSE for SNR} and \ref{table:HVI and IAT}.
The baselines in Table~\ref{table:HVI and IAT} include HVI-CIDNET~\cite{yan2025hvi} and IAT~\cite{Cui_2022_BMVC}.

Moreover, we note that inserting our strategy into HVI-CIDNet results in a decrease of 16.1\% in SSIM and 4.1\% in LPIPS on SMID. 
We explain this phenomenon as a result of using the same hyper-parameter $\lambda$ in Eq~\ref{eq:final} across different datasets, showing the robustness of our method across diverse scenes.
However, this hyper-parameter can be adjusted for different methods to achieve better results. For HVI-CIDNet, which includes an additional loss term in the HVS space alongside the RGB space, keeping $\lambda$ unchanged may relatively overweight this term and diminish the impact of the depth supervision loss $\mathcal{L}_d$ in Eq~\ref{eq:final}, leading to poorer depth estimation on low-light data. By increasing $\lambda$, we could observe improved performance that surpasses the baseline in all metrics.

Furthermore, we compare the optimal results of our method with a lot of representative low-light enhancement methods, including KinD, URetinex-Net++, LLFlow-L-SKF, Retinexmamba, Wakeup-Darkness, Zero-DCE, RUAS, SCI, NeRCO, and Clip-LIE. These methods contain the strategies of both supervised and unsupervised. The results are exhibited in Table~\ref{table:quantitative4}, which demonstrate the SOTA performance brought by our strategy.

\begin{table*}[t]
\centering
\caption{The comparison between our method and several representative low-light image enhancement baselines.}
\label{table:quantitative4}
\resizebox{\textwidth}{!}{
\begin{tabular}{l|p{1cm}<{\centering}p{1cm}<{\centering}p{1cm}<{\centering}p{1cm}<{\centering}|p{1cm}<{\centering}p{1cm}<{\centering}p{1cm}<{\centering}p{1cm}<{\centering}}
	\toprule
	& \multicolumn{4}{c|}{SDSD-indoor} &\multicolumn{4}{c}{SDSD-outdoor} \\
	\cline{1-9}
	Methods & PSNR & SSIM& LPIPS & RMSE & PSNR & SSIM& LPIPS & RMSE \\
	\hline
	KinD(Supervised)\cite{zhang2019kindling}
	&19.13 &0.876 &0.346 &53.462&18.10 &0.761 &0.346 &28.540  \\
        URetinex-Net++(Supervised)\cite{wu2025interpretable}
	&20.29 &0.922 &0.204 &46.030&17.70 &0.520 &0.265 &47.800 \\
	LLFlow-L-SKF(Supervised)\cite{wu2023skf}
	&21.58 &0.980 &0.275 &40.260&24.89 &0.872 &0.423 &22.085  \\
        Retinexmamba(Supervised)\cite{bai2024retinexmamba}
	&22.84 &0.985 &0.015 &7.490&18.56 &0.939 &0.072 &15.860 \\
    \hline
        Wakeup-Darkness(unsupervised)\cite{zhang2025}
        &18.51 &0.714 &0.413 &46.032&18.43 &0.710 &0.365 &36.256 \\
        
        Zero-DCE(Unsupervised)\cite{Zero-DCE}
	&17.81 &0.787 &0.361 &58.046&17.70 &0.520 &0.265 &47.796  \\
        RUAS(Unsupervised)\cite{liu2021ruas}
        &13.97 &0.717 &0.467 &82.992&11.19 &0.469 &0.307 &92.194\\
        SCI(Unsupervised)\cite{ma2022toward}
        &17.54 &0.872 &0.389 &60.102&21.68 &0.783 &0.283 &32.570\\
        NeRCO(Unsupervised)\cite{Yang_2023_ICCV}
        &21.73 &0.849 &0.185 &41.130&22.25 &0.843 &0.334 &38.261\\
        Clip-LIE(Unsupervised)\cite{liang2023iterative}
        &16.72 &0.852 &0.393 &64.417&20.59 &0.862 &0.318 &34.492\\
	\hline
	Ours+SNR &\textbf{26.86} &\textbf{0.924} &\textbf{0.063}&\textbf{13.028} &\textbf{29.38} &\textbf{0.858}&\textbf{0.135}&\textbf{9.429} \\
	\bottomrule[1pt]
\end{tabular}}
\end{table*}

\begin{table*}[tb]
    \caption{The quantitative comparison illustrates the range of plausible outcomes for methods with and without our strategy. We report the average variance value here.
    Incorporating our strategy reduces the variance among outputs from different low-light enhancement methods under the same supervision, thereby alleviating the ill-posed nature of the problem.}
    \label{table:quantitative44}
    \renewcommand\arraystretch{1.15}
    \centering
    \resizebox{\textwidth}{!}{
    \begin{tabular}{l|p{2cm}<{\centering}|p{2cm}<{\centering}|p{2cm}<{\centering}|p{2cm}<{\centering}|p{2cm}<{\centering}|p{2cm}<{\centering}}
		\toprule
	Methods&\multicolumn{1}{c|}{SDSD-indoor} &
			\multicolumn{1}{c|}{SDSD-outdoor} & 
			\multicolumn{1}{c|}{SMID} & 
			\multicolumn{1}{c|}{LOLv2-real} & 
			\multicolumn{1}{c|}{LOLv2-synthetic}& 
			\multicolumn{1}{c}{SID}\\
        \cline{1-7}
		Baseline &1.95$\times 10^{-2}$ &7.62$\times 10^{-3}$ &5.72$\times 10^{-3}$ &3.06$\times 10^{-2}$ &2.49$\times 10^{-2}$ &2.92$\times 10^{-2}$ \\
            Ours &\textbf{4.24$\times 10^{-3}$} &\textbf{5.09$\times 10^{-3}$} &\textbf{4.58$\times 10^{-3}$} &\textbf{2.34$\times 10^{-2}$}  &\textbf{2.09$\times 10^{-2}$} &\textbf{6.03$\times 10^{-3}$} \\
			\bottomrule 
\end{tabular}}

\vspace{0.1in}
\caption{The average $L_2$ distance between the features of low-light and normal-light images reflects their alignment potential, demonstrating the effectiveness of our learned features.}
    \label{table:quantitative55}
    \renewcommand\arraystretch{1.15}
    \centering
    \resizebox{\textwidth}{!}{
    \begin{tabular}{l|p{2cm}<{\centering}|p{2cm}<{\centering}|p{2cm}<{\centering}|p{2cm}<{\centering}|p{2cm}<{\centering}|p{2cm}<{\centering}}
		\toprule
	Methods&\multicolumn{1}{c|}{SDSD-indoor} &
			\multicolumn{1}{c|}{SDSD-outdoor} & 
			\multicolumn{1}{c|}{SMID} & 
			\multicolumn{1}{c|}{LOLv2-real} & 
			\multicolumn{1}{c|}{LOLv2-synthetic}& 
			\multicolumn{1}{c}{SID}\\
        \cline{1-7}
		SNR &8.87$\times 10^3$ &3.97$\times 10^3$ &7.33$\times 10^3$ & 1.16$\times 10^4$&8.15$\times 10^3$ &2.56$\times 10^3$ \\
            +Ours &\textbf{4.89$\times 10^3$} &\textbf{1.79$\times 10^3$} &\textbf{4.24$\times 10^3$} &\textbf{5.02$\times 10^3$}  &\textbf{5.94$\times 10^3$} &\textbf{2.21$\times 10^3$} \\
            \hline
            Restormer &9.18$\times 10^3$ &1.25$\times 10^3$ &9.51$\times 10^3$ &7.32$\times 10^3$ &4.18$\times 10^3$ &6.07$\times 10^2$ \\
            +Ours &\textbf{6.08$\times 10^1$} &\textbf{3.11$\times 10^2$} &\textbf{1.81$\times 10^2$} &\textbf{3.68$\times 10^3$}  &\textbf{1.83$\times 10^1$} &\textbf{6.04$\times 10^2$}\\
			\bottomrule
\end{tabular}}
\end{table*}

\begin{figure*}[tb]
    \centering
    \includegraphics[width=1\linewidth]{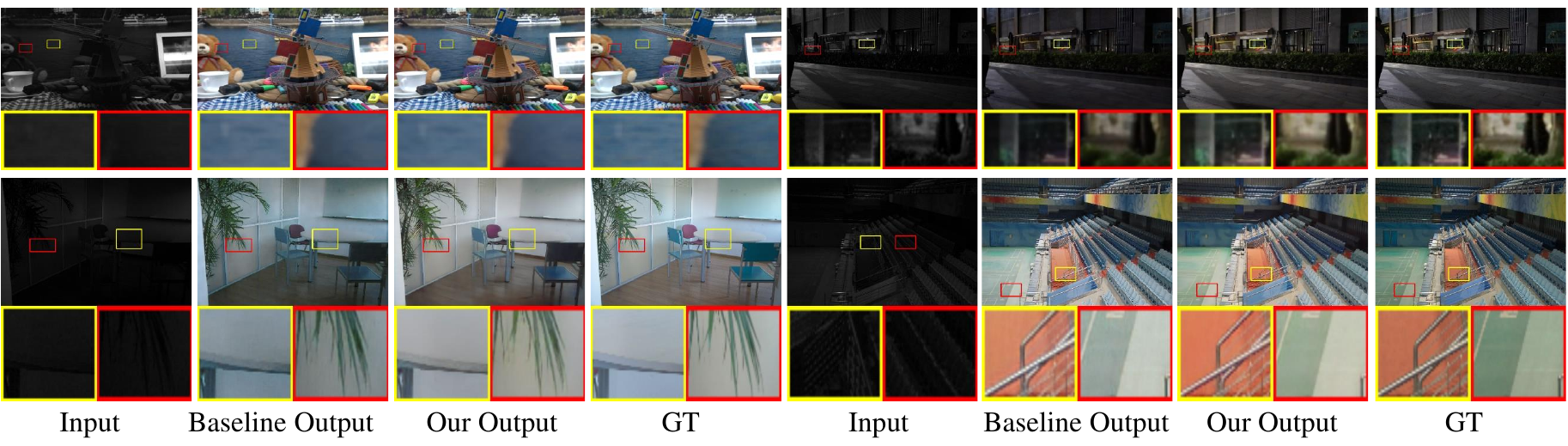}
    \caption{Visual comparison on different datasets. From top to down, they are Restormer in SMID and SDSD-outdoor, and SNR in LOLv2-real. Our method yields better results.}
    \vspace{-0.1in}
    \label{fig:Quantitative Evaluation1}
\end{figure*}

\begin{figure*}[tb]
    \centering
    \includegraphics[width=1\linewidth]{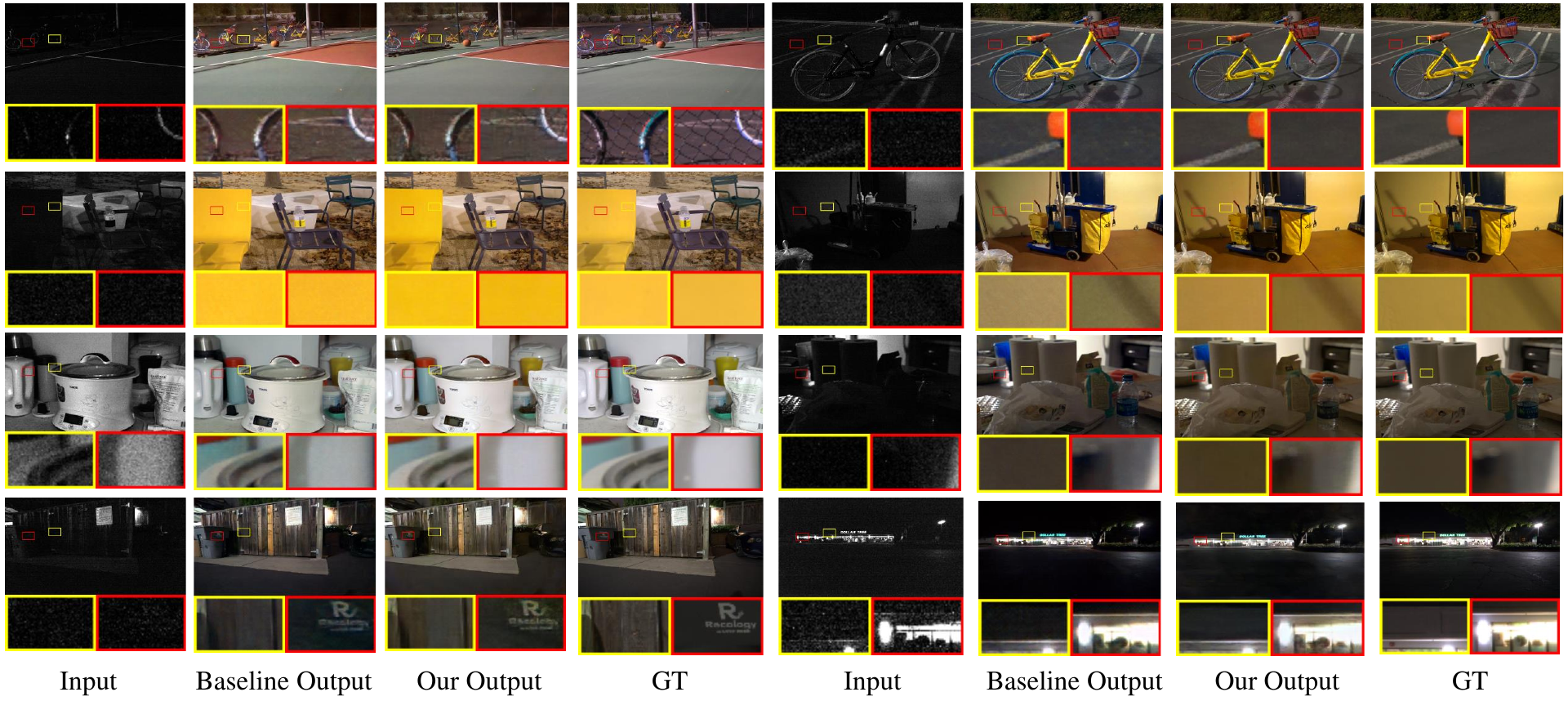}
    \vspace{-0.1in}
    \caption{Visual comparisons of Restormer in the SID dataset.}
    \vspace{-0.1in}
    \label{fig:3}
\end{figure*}

\begin{figure*}[h!]
    \centering
    \includegraphics[width=1.0\linewidth]{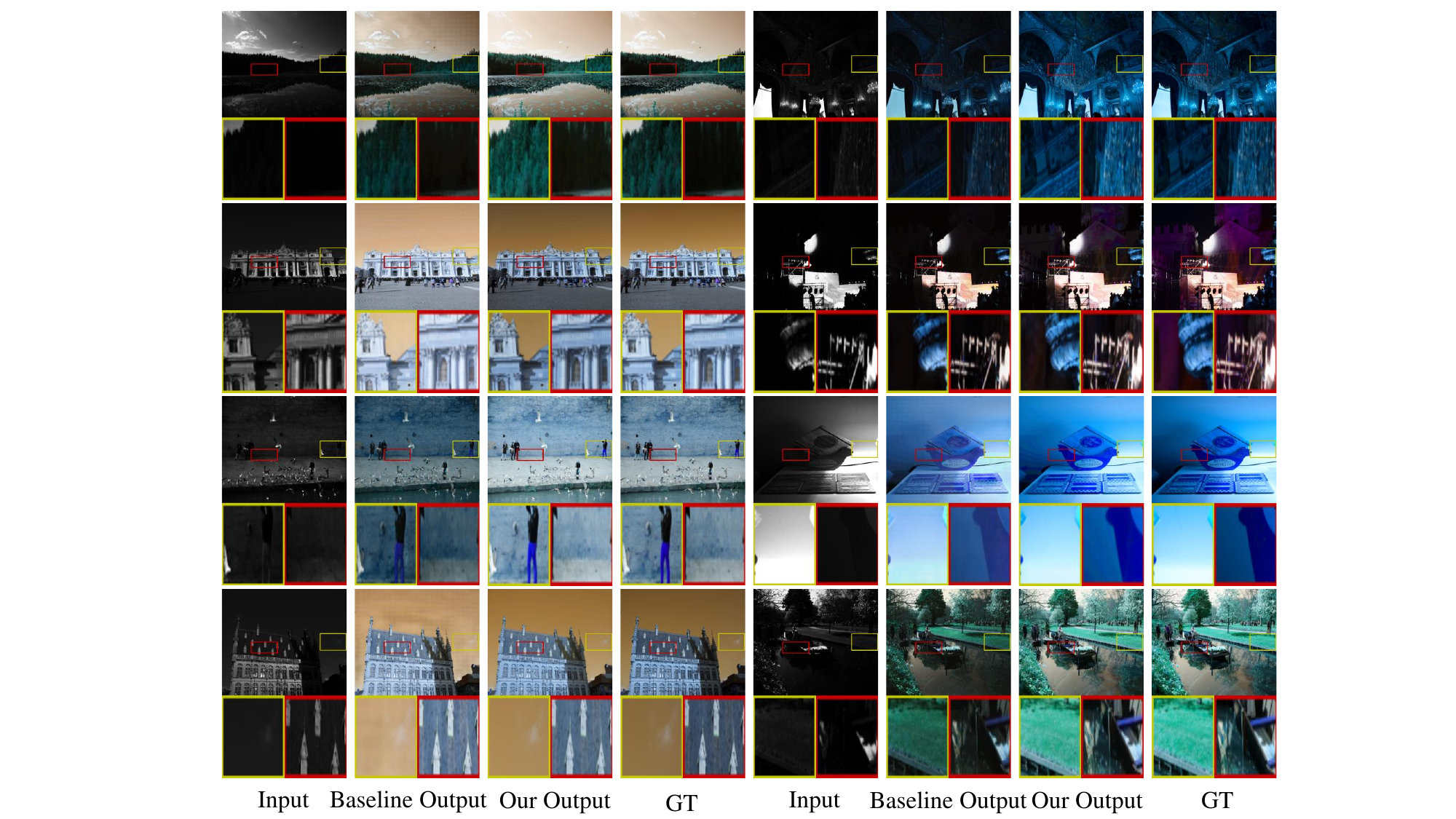}
    \vspace{-0.1in}
    \caption{Visual comparisons of SNR in the LOLV2-synthetic dataset.}
    \label{fig:4}
    \vspace{-0.1in}
\end{figure*}

\begin{figure*}[h!]
    \centering
    \includegraphics[width=1.0\linewidth]{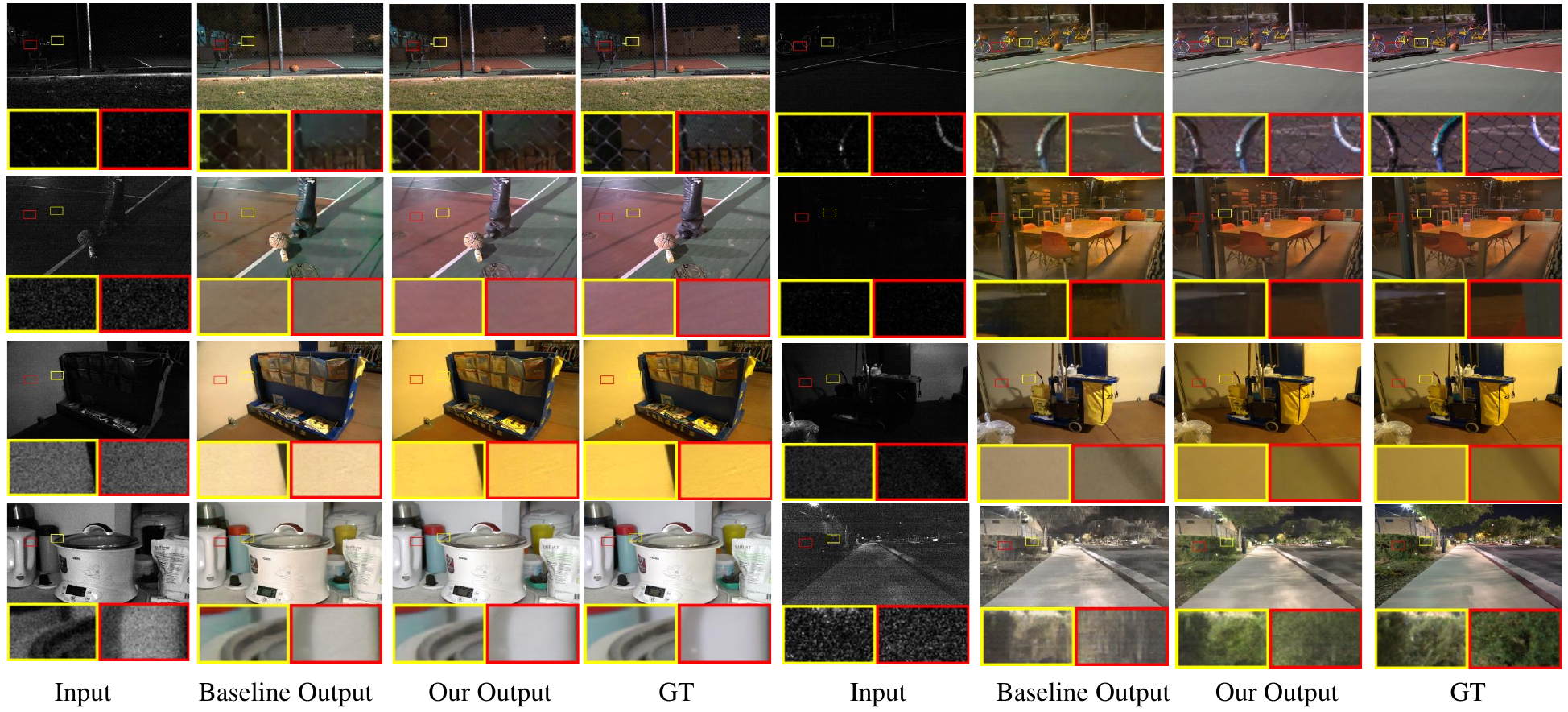}
    \vspace{-0.1in}
    \caption{Visual comparisons of SNR in the SID dataset.}
    \label{fig:5}
    \vspace{-0.1in}
\end{figure*}

\vspace{1mm}
\noindent\textbf{Evaluation for the range of plausible outcomes.}
As stated in the introduction, we think that incorporating priors imposes additional constraints on the enhancement process, thereby reducing the range of plausible outcomes. 
This statement highlights that low-light image enhancement is a highly ill-posed problem, where even under identical supervision (i.e., the same low-light and normal-light data pairs), the outputs from different methods can vary significantly while remaining plausible.
To verify this claim, we conduct a corresponding evaluation.
We compute the distances between outputs of different LLIE methods and the corresponding ground truths. These LLIE methods include Retinexformer, Restormer, and SNR. For each image in the evaluation datasets, this yields a triplet representing the variation in ``output deviation from ground truth" achievable by LLIE models. The variance of these triplets reflects the range of plausible outcomes for the dataset, with lower variance indicating a narrower range.
As shown in the Table~\ref{table:quantitative44}, we perform this evaluation across different datasets and compute the average variance of the triplets for all images within each dataset. The results demonstrate that the range of plausible outcomes produced by our method with priors is significantly narrower than that of the baselines, supporting our claim.

\vspace{1mm}
\noindent\textbf{Analyses: our features are indeed beneficial?}
We have designed an experiment to evaluate whether the deep features in our method are effectively benefited.
Since our fusion of image features and depth-aware features occurs in the encoder, we focus on analyzing the features in the deepest layer. Specifically, we feed low-light data into the network to obtain ``feature A" and normal-light data to obtain ``feature B" (to this end, we finetune the network using normal-light input data with a reconstruction loss).
The distance between these features reflects the distribution deviation between low-light and normal-light data in the network’s deep feature space. Since normal-light features are more easily aligned with the ground truth, a smaller distance between ``feature A" and ``feature B" indicates more beneficial features for improved alignment.
As shown in the Table~\ref{table:quantitative55}, our method yields smaller distances, demonstrating that the enhanced features are indeed beneficial.

\begin{figure*}[tb]
    \centering
    \includegraphics[width=1\linewidth]{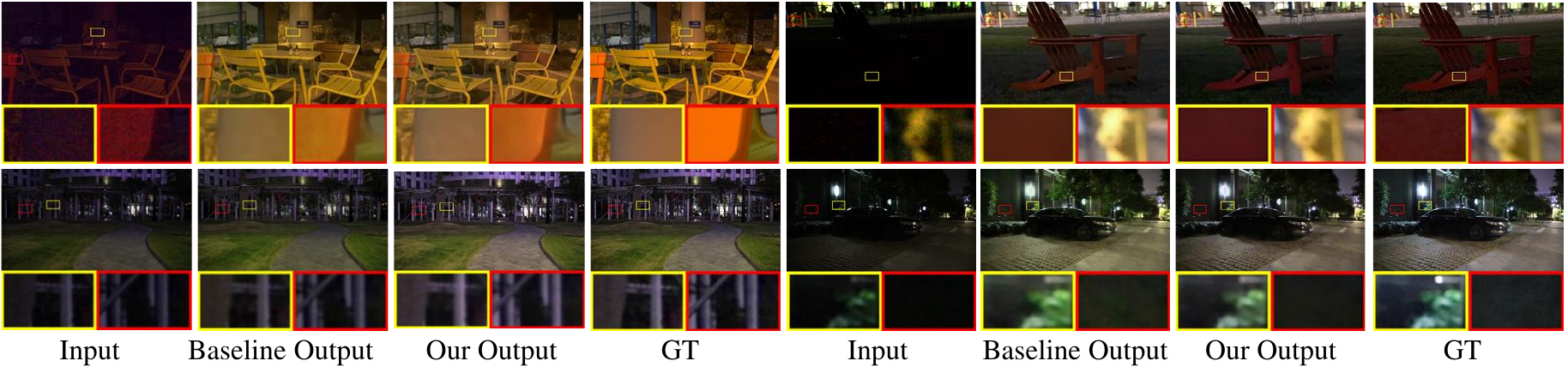}
    \vspace{-0.15in}
    \caption{Visual comparisons of Retinexformer in the SID (top) and SDSD-outdoor dataset (bottom).}
    \label{fig:Quantitative Evaluation2}
    \vspace{-0.1in}
\end{figure*}

\vspace{1mm}
\noindent\textbf{Qualitative evaluation.}
We also provide the visual comparisons for the baseline and ours on different datasets, as shown in Figs.~\ref{fig:Quantitative Evaluation1}, \ref{fig:3}, \ref{fig:4}, \ref{fig:5}, and \ref{fig:Quantitative Evaluation2}.
As we can see, our method's results are closer to the ground truth with natural illumination and color, fewer noise and artifacts. Our approach offers several well-founded advantages: the depth information provides additional 3D cues, enabling us to capture the actual lighting conditions that conform to physical laws. Consequently, the depth features can contribute to color accuracy, sharper edge boundaries, and better foreground/background separation.

\subsection{Ablation Study}
\label{sec:ab}
In this section, we perform various ablation studies on the SDSD datasets, encompassing both indoor and outdoor subsets, to assess the significance of different components in our framework. Specifically, we evaluate the key components in HDGFFM through several kinds of ablated cases:
(i) The location where the geometric factor is added.
(ii) The method of fusion between image features and depth-aware features.
(iii) Different weights for depth loss. 
(iv) The influence of the choice on pre-trained depth estimator. (v) Others.

\begin{itemize}
\item[$\bullet$]\textbf{``Ours with Fusion in Decoder''}. 
As indicated in Sec.~\ref{sec:overview} and Eq~\ref{eq:encoder}, the fusion in our framework occurs in the encoder of the target network $\mathcal{G}$. This choice is made because both the image content features and depth-aware features are extracted from the low-light input image, fostering homogeneity and enhancing fusion effects. To investigate the impact of alternative fusion points, we aim to test whether fusing image content features and depth-aware features in the decoder of $\mathcal{G}$ yields comparable results.

\item \textbf{``Ours w/o $w_l$''}. 
In this ablation setting, we omit the correlation computation step (i.e., $w_l$ in Eq~\ref{eq:sigmoid}), and the fusion procedure (Eq~\ref{eq:fusion-eq}) can be written as $\bar{f}_{g,l}= f''_{g,l} + f'_{g,l}+f_{g,l}$.

\item[$\bullet$]\textbf{``Ours with Alternative Attention''}. 
In this ablation study, we explore the impact of different attention map formulation strategies for acquiring information from depth-aware features.
In other words, we compute the attention map using only the depth-aware features, and employ the original image features to predict the value part: replacing $W_q(f_{g,l})$ in Eq~\ref{eq:cross-attention} as $\hat{W}_q(\hat{f}_{d,l})$ and $\hat{W}_v(\hat{f}_{d,l})$ as $\hat{W}_v(f_{g,l})$.

\item[$\bullet$]\textbf{``Ours with $\lambda \times 5$'' and ``Ours with $\lambda \div 5$''}.
As specified in Eq~\ref{eq:final}, there is a hyper-parameter that governs the contribution of image reconstruction loss and depth prediction loss, denoted as $\lambda$. We aim to analyze the effects of varying the value of $\lambda$ on the final results.

\item[$\bullet$]\textbf{``Ours with Depth Anything'' and ``Ours with Depth Anything2''}. 
We integrate a new SOTA depth estimation approach, Depth Anything~\cite{yang2024depthany} and Depth Anything V2~\cite{yang2024depth2}, into our framework to replace DPT, demonstrating that our method's performance is not dependent on the choice of the pre-trained $\mathcal{D}$.

\item[$\bullet$]\textbf{``Depth Feature I'' and ``Depth Feature II''}. 
In this ablation setting, we study whether our depth-aware image feature is better than depth features extracted from depth maps. Thus, we set two ablation experiments. ``Depth Feature I" means extracting depth features directly from low-light images' depth maps; ``Depth Feature II" means obtaining depth features from initially enhanced low-light images' depth maps.

\item[$\bullet$]\textbf{``NYU/KITTI finetune''}. 
We aim to explore the impact of using datasets with real-captured depth, such as NYU or KITTI, and synthesizing corresponding low-light and normal-light pairs. The experimental results are obtained by finetuning a model with these additional NYU/KITTI samples.

\item[$\bullet$]\textbf{``depth estimator params+''}. 
We aim to analyze whether the performance of our framework can be enhanced with the enhanced model capability of the depth estimator $\mathcal{F}$. We conduct an experiment by increasing the model capability of the depth estimator, using a transformer.

\end{itemize}

\begin{table*}[t]
\centering
\caption{Results of the ablation study in the SDSD-indoor, SDSD-outdoor and SMID datasets with DP3DF. Our method performs better than all ablation settings.}
\label{comparison-abla}
\resizebox{\textwidth}{!}{
\begin{tabular}{l|p{1cm}<{\centering}p{1.5cm}<{\centering}|p{1.5cm}<{\centering}p{1.5cm}<{\centering}|p{1.5cm}<{\centering}p{1.5cm}<{\centering}p{1.5cm}<{\centering}p{1.5cm}<{\centering}p{1.5cm}<{\centering}p{1.5cm}<{\centering}p{1.5cm}<{\centering}}
	\toprule
	& \multicolumn{2}{c|}{SDSD-indoor} &\multicolumn{2}{c|}{SDSD-outdoor}&\multicolumn{2}{c}{SMID}   \\
	\cline{1-7}
	Methods & PSNR & SSIM& PSNR & SSIM & PSNR & SSIM \\
	\hline
	Ours with Fusion in Decoder
	&28.97 &0.875 &27.34 &0.799 &24.83 &0.705 \\
        Ours w/o $w_l$
	&30.75 &0.895 &27.07 &0.775 &26.38 &0.746 \\
        Ours with Alternative Attention
	&30.52 &0.892 &27.98 &0.804 &26.18 &0.740  \\
	Ours with $\lambda\times5$
	&29.53 &0.878 &\textbf{27.99} &0.802 &\textbf{26.43} &\textbf{0.747} \\
        Ours with $\lambda\div 5$
	&29.89 &0.883 &27.42 &0.795 &26.31 &0.744  \\
        Ours with Depth Anything
	&30.82 &0.895 &27.66 &0.801 &26.28 &0.742  \\
        Ours with Depth Anything V2
        &30.53&0.891&27.89&0.805&26.22&0.742\\
        Depth Feature I
        &30.77&0.894&26.52&0.789&26.05&0.734\\
        Depth Feature II
        &30.68&0.893&27.10&0.794&26.65&0.750\\
        NYU/KITTI fine-tune
        &30.75&0.895&27.69&0.802&25.68&0.728\\
        depth estimator params+
        &30.66&0.892&27.47&0.799&25.91&0.737\\
	\hline
	Ours &\textbf{30.95} &\textbf{0.897} &\textbf{27.89}&\textbf{0.804} &\textbf{26.35} &\textbf{0.743} \\
	\bottomrule[1pt]
\end{tabular}}
\vspace{-0.1in}
\end{table*}        

\begin{table*}
\centering
\caption{Results of the ablation study in the SDSD-indoor, SDSD-outdoor and SMID with SNR.
Our method performs better than all ablation settings.}
\label{comparison-abla2}
\resizebox{\textwidth}{!}{
\begin{tabular}{l|p{1cm}<{\centering}p{1.5cm}<{\centering}|p{1.5cm}<{\centering}p{1.5cm}<{\centering}|p{1.5cm}<{\centering}p{1.5cm}<{\centering}p{1.5cm}<{\centering}p{1.5cm}<{\centering}p{1.5cm}<{\centering}p{1.5cm}<{\centering}p{1.5cm}<{\centering}}
	\toprule
	& \multicolumn{2}{c|}{SDSD-indoor} &\multicolumn{2}{c|}{SDSD-outdoor}&\multicolumn{2}{c}{SMID}  \\
	\cline{1-7}
	Methods & PSNR & SSIM& PSNR & SSIM& PSNR & SSIM\\
	\hline
	Ours with Fusion in Decoder
	&23.50 &0.877 &29.14 &0.836 &28.50 &0.806\\
        Ours w/o $w_l$
	&22.40 &0.868 &18.76 &0.706 &28.70 &0.811 \\
        Ours with Alternative Attention
	&21.75 &0.857 &29.64 &0.846 &28.56 &0.809 \\
	Ours with $\lambda\times5$
	&23.92 &0.900 &\textbf{30.00} &0.853 &28.65 &0.813  \\
        Ours with $\lambda\div 5$
	&23.75 &0.896 &\textbf{30.03} &0.851 &28.69 &0.813 \\
     Ours with Depth Anything
	&\textbf{27.18} &\textbf{0.930} &28.47 &0.846 &28.78 &\textbf{0.851}  \\
        Ours with Depth Anything V2
        &\textbf{26.91}&\textbf{0.933}&29.03&0.848&28.72&0.812
    \\
        Depth Feature I
        &26.68&0.923&28.48&0.850&28.49&0.811
    \\
        Depth Feature II
        &27.70&0.926&28.98&0.851&28.73&0.819
    \\
        NYU/KITTI fine-tune
        &25.44&0.916&29.30&0.855&28.22&0.804
    \\
        depth estimator params+
        &25.07&0.912&28.75&0.850&28.79&0.812
    \\
	\hline
	Ours &\textbf{26.86} &\textbf{0.924} &\textbf{29.38}&\textbf{0.858} &\textbf{29.01} &\textbf{0.813}\\ 
	\bottomrule[1pt]
\end{tabular}}

\end{table*}

\begin{table}
\centering
\caption{Depth estimation evaluation: results of RMSE between $\hat{d}_d$ and $d_n$ in Eq~\ref{eq:depth loss} when employing DP3DF. }
\label{depth metrics}
\resizebox{1.0\linewidth}{!}{
\begin{tabular}{l|c|c|c}
	\toprule
	\multirow{1}{*}{Methods}& \multicolumn{1}{c|}{SDSD-indoor}& \multicolumn{1}{c|}{SDSD-outdoor}
        & \multicolumn{1}{c}{SMID}\\
	\cline{2-4}
	\hline
	DPT
	&0.034&0.039&0.072 \\
	Depth Anything
	&0.007&0.011&0.009 \\
        Depth Anything V2
	&0.006&0.007&0.014 \\
        \bottomrule 
\end{tabular}}
\vspace{-0.1in}
\end{table}

As depicted in Tables.~\ref{comparison-abla} and \ref{comparison-abla2}, it is evident that the results of all ablation settings are inferior to our original implementation.
In comparing ``Ours'' with ``Ours with Fusion in Decoder'', we validate the impracticality of fusing restored image content features and depth-aware features since the former is close to the normal-light output while the latter is close to the low-light inputs. Furthermore, comparing ``Ours'' with ``Ours w/o $w_l$.'' and ``Ours with Alternative Attention'' underscores the significance of our proposed ``Correlation-based Cross Attention for Fusion'' strategy. Lastly, the consistent and even improved results across different values for $\lambda$ highlight the robustness of the hyperparameter choice in our framework. This emphasizes that researchers can choose optimal $\lambda$ values for different datasets.

Additionally, the ablation results of ``Ours with Depth Anything" and ``Ours with Depth Anything V2" demonstrate that our method's performance is not dependent on the choice of the pre-trained depth estimation network. This is because, as shown in Table.~\ref{depth metrics}, our method can accurately predict depth values for low-light inputs with minimal error under the supervision provided by both DPT, Depth Anything, and Depth Anything V2.

Moreover, the comparisons among ``Depth Feature I", ``Depth Feature II", and ``Ours" support that the depth-aware image features are better than depth features extracted from the depth maps in most cases. This is because depth-aware image features and low-light image features are homologous, which are more suitable for fusion. 
We should also note that extracting predicting depth maps directly from the low-light images will face the issue of domain shift. Also, there is a gap between the output space of the initial low-light image enhancement method and the input scope of the subsequent depth estimator, leading to depth prediction errors.

Meanwhile, comparing ``NYU/KITTI fine-tune" and ``Ours", we can see that the performance with NYU and KITTI data is not superior. This may due to domain gaps between NYU/KITTI and low-light image enhancement datasets like SDSD. Furthermore, the predictions from the state-of-the-art depth estimator are generally good, not very worse than the ground truth. This is particularly true for real-world datasets like KITTI, where annotations are often sparse.
Furthermore, increasing the model capability of the depth estimator may not improve the results, by comparing ``depth estimator params+" and ``Ours". This might because the lack of sufficient data to train the larger model, and the ground truth employed is not changed compared with the small model.

\noindent\textbf{Which is the source of performance improvement, depth information or additional feature space?}
In the setting of ``Ours with decoder'', the fusion is conducted in the decoder part, where fusion is problematic since these two features are heterogeneous (they are close to the image and depth outputs, respectively). In this situation, although the additional feature space of the depth estimator is utilized, the performance improvement is not substantial. Thus, the improvement does not primarily stem from the addition of the feature space. On the other hand, if the fusion occurs in the encoder part, the improvement is noticeable. Consequently, depth information is a helpful source, but it requires a suitable strategy for fusion.

\begin{figure}[!t]
	\begin{center} 
		\includegraphics[width=1.2\linewidth]{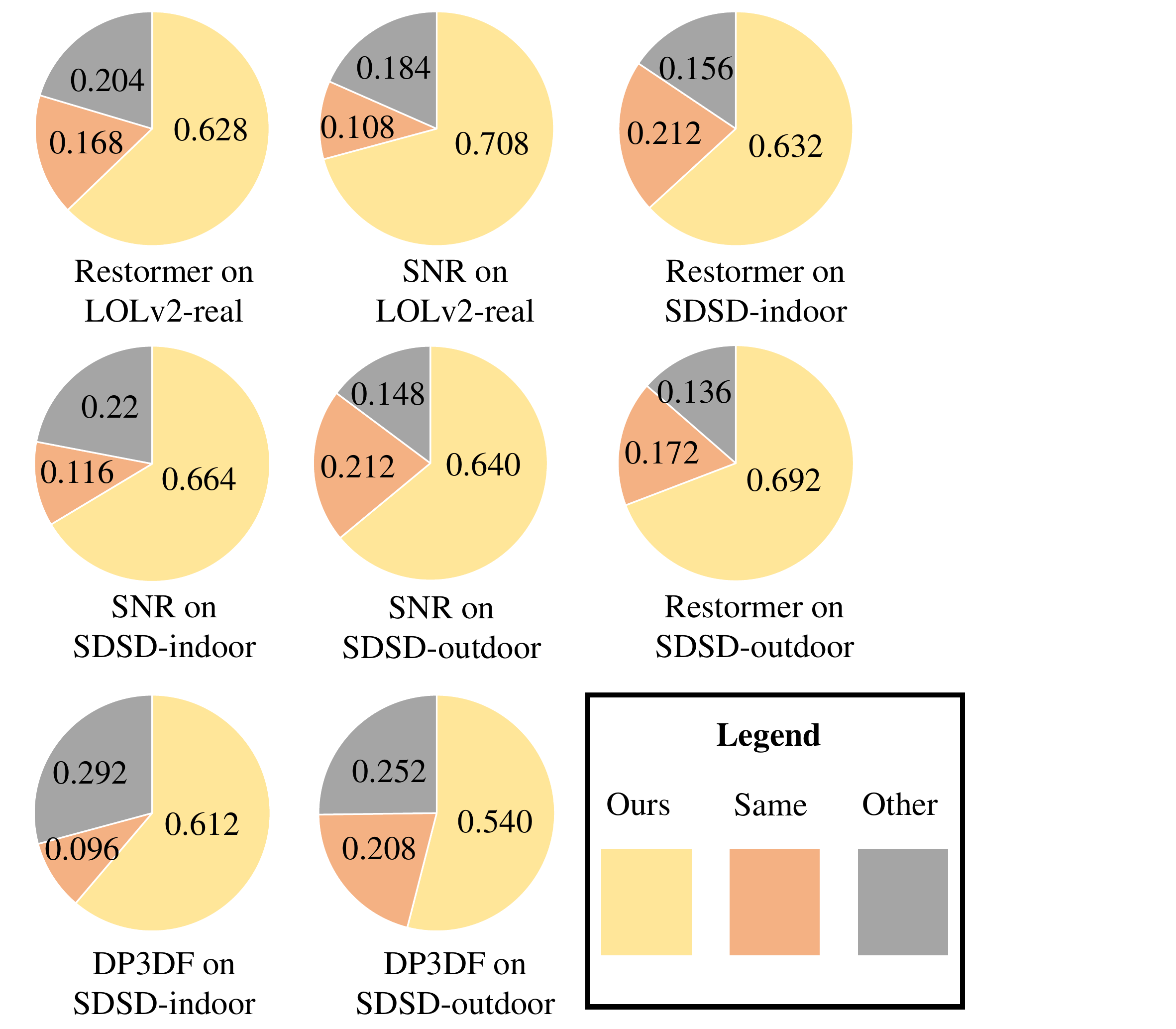}
        \vspace{-0.1in}
	\end{center}
	\caption{
		The results of the user study, which summarize that the results enhanced with our strategy are preferred by participants compared with the baselines' results.
	}
	\label{us_tbl}
\end{figure}

\subsection{User Study}
To evaluate the effectiveness of our proposed HDGFFM in terms of subjective evaluation, we carried out a user study with 50 participants (using online questionnaires). Our objective is to verify the subjective quality of our approach compared with baselines, and the evaluation scenarios should cover across various datasets and frameworks.

The results of the user study are presented in Fig.~\ref{us_tbl}. Clearly, the participants consistently preferred the results of our method. This indicates that our approach significantly improves the human subjective perception of existing frameworks.

\subsection{Failing Cases}
\begin{figure*}
    \centering
    \includegraphics[width=1\linewidth]{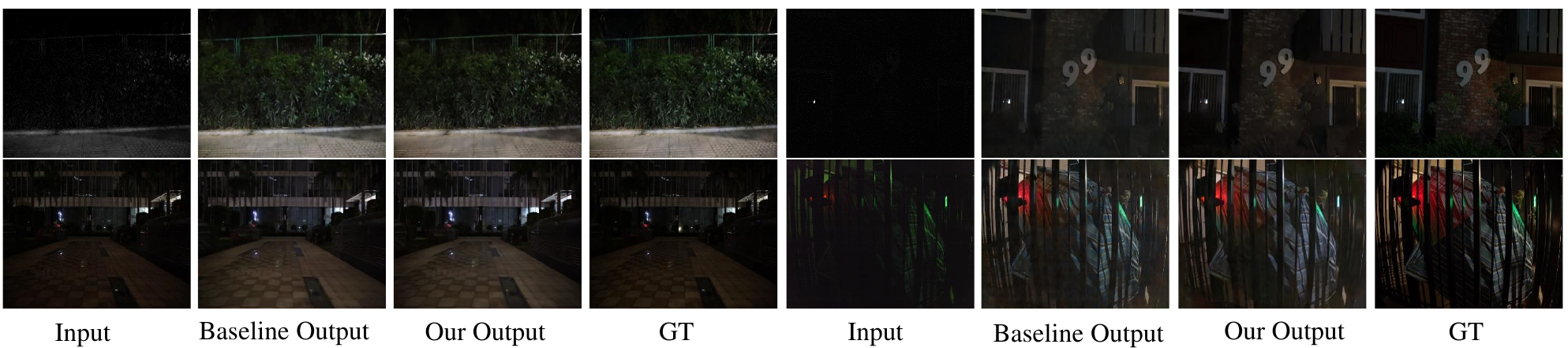}
    \caption{Failure cases. The first row shows results using SNR: the left image corresponds to SDSD-outdoor, and the right to SMID. The second row presents results using DP3DF: again, the left image is from SDSD-outdoor, and the right from SMID.}
    \label{fig:Failing Case2}
\end{figure*}

We provided several failure cases as shown in Fig.~\ref{fig:Failing Case2}. As seen in these images, the performance of the baselines and their versions with our method are similar. These areas mainly consist of images with complex textures, where depth information is difficult to predict. As a result, the model pays less attention to utilizing the depth information in these regions. Thus, the depth priors are beneficial in areas where depth information is accurately predicted. In these regions, the depth information can provide clear boundaries between different objects, enhancing sharp boundaries and improving foreground/background separation.

We also perform several quantitative analyses to support our hypothesis. 
We compute the depth prediction error for each pixel location in observed failure cases by measuring the distance to the ground truth. We choose the results with DPT for experiments.
As shown in Table~\ref{depth metrics2}, we observe that the overall distribution of these errors is higher than the average value. This aligns with the observation that the improvement on these samples is weaker, as baseline results and those enhanced with our method are more similar, as shown in Fig.~\ref{fig:Failing Case2}.
Moreover, we analyze the distribution of $w_l$ in Eq~\ref{eq:sigmoid} for these samples, as it represents the contribution of depth-aware features to the network backbone after the cross-attention computation. 
As shown in Table~\ref{weighting-factor}, we find that $w_l$ values are lower for these failure cases, compared to other samples. 
These results indicate that depth-aware features are not effectively learned, and the model tends to underutilize them in these regions.

\begin{table}
\centering
\caption{Depth estimation evaluation: results of RMSE between $\hat{d}_d$ and $d_n$ in Eq~\ref{eq:depth loss} with DP3DF. A comparison between failure cases and other results is shown here.}
\label{depth metrics2}
\resizebox{1.0\linewidth}{!}{
\begin{tabular}{l|p{2.2cm}<{\centering}|p{2.2cm}<{\centering}}
	\toprule
	\multirow{1}{*}{Methods}& \multicolumn{1}{c|}{SDSD-outdoor}
        & \multicolumn{1}{c}{SMID}\\
	\cline{2-3}
	\hline
	DPT
	&0.039&0.072 \\
        Failing Case
        &0.127&0.089\\
        \bottomrule 
\end{tabular}}
\vspace{0.1in}

\caption{Average value of $w_l$ in Eq~\ref{eq:sigmoid} and \ref{eq:fusion-eq} reflects the contribution of the depth-aware feature to final fused feature.}
\label{weighting-factor}
\resizebox{1.0\linewidth}{!}{
\begin{tabular}{l|p{2.2cm}<{\centering}|p{2.2cm}<{\centering}}
	\toprule
	\multirow{1}{*}{Methods}& \multicolumn{1}{c|}{SDSD-outdoor}
        & \multicolumn{1}{c}{SMID}\\
	\cline{2-3}
	\hline
	Normal Case
	&0.51&0.50 \\
        Failing Case
        &0.40&0.38\\
        \bottomrule 
\end{tabular}}
\end{table}

\section{Limitation and Future Work}
In this paper, we introduce a novel emphasis on employing geometrical priors, particularly in the form of depth, for low-light enhancement, an aspect overlooked by previous methods. Our extensive experiments across various datasets and networks demonstrate the effectiveness of our proposed HDGFFM. Despite achieving significant advancements, there are still areas for improvement. Firstly, further efforts are required to minimize model parameters and additional computational requirements of HDGFFM, even though it has not introduced heavy computations in this paper. Secondly, we plan to extend our strategy to incorporate geometrical priors for additional low-level vision tasks, aiming to develop a unified and generalizable theory.

Additionally, we plan to investigate the use of other types of geometric information that could potentially enhance the performance of LLE tasks.
For example, we believe our method can be extended to surface normals. 
This is because there are state-of-the-art surface normal prediction methods that can be applied to normal-light images to obtain the ground truth.

\section{Conclusion}
This paper presents the first attempt to leverage geometric priors (specifically depth information) for low-light enhancement, covering both image and video enhancement tasks.
Rather than extracting features directly from depth maps, we propose to derive depth-aware features from low-light images, facilitating easier fusion with image features. To achieve this, we introduce a network that predicts depth information from an input low-light image. This lightweight depth estimator distills depth estimation capabilities from large-scale, open-world depth models. Furthermore, the extracted depth-aware features are efficiently fused with the encoder features of the target network using our proposed HDGFFM module, which incorporates correlation-based cross-attention to combine the strengths of both self-attention and cross-attention mechanisms. Extensive experiments on representative datasets and methods validate the effectiveness of our proposed approach.

\bibliographystyle{IEEEtran}
\bibliography{main}
\newpage 

\begin{IEEEbiography}[{\includegraphics[width=1in,height=1.3in,clip]{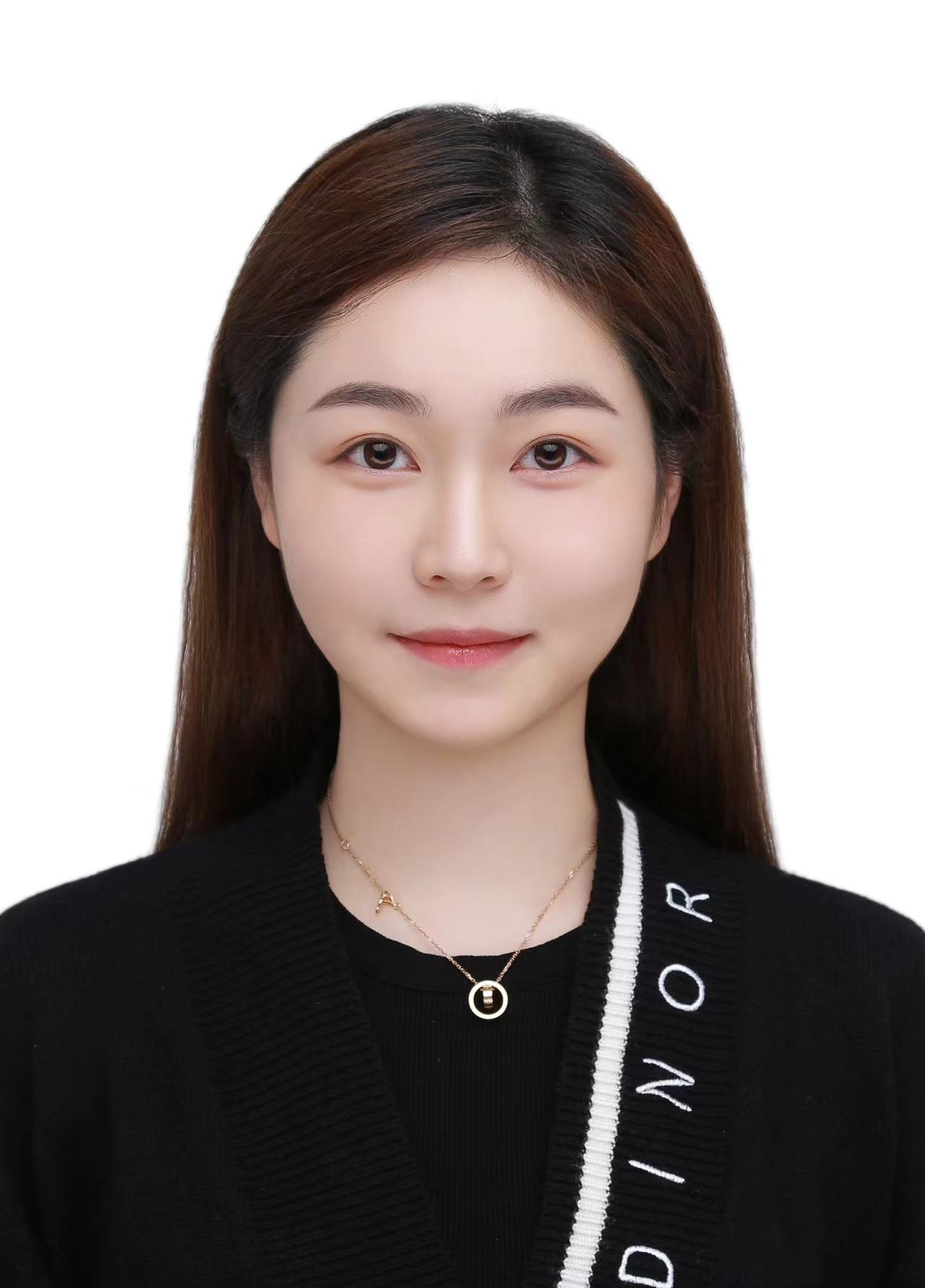}}]{Yingqi Lin} received her bachelor degree from Shih Hsin University, and received master degree from University of Southern California in 2021. She is currently working at Zhejiang Lab, where her research focuses on computer vision, image processing, and multimodal models,etc.
\end{IEEEbiography}

\begin{IEEEbiography}[{\includegraphics[width=1in,height=1.3in,clip]{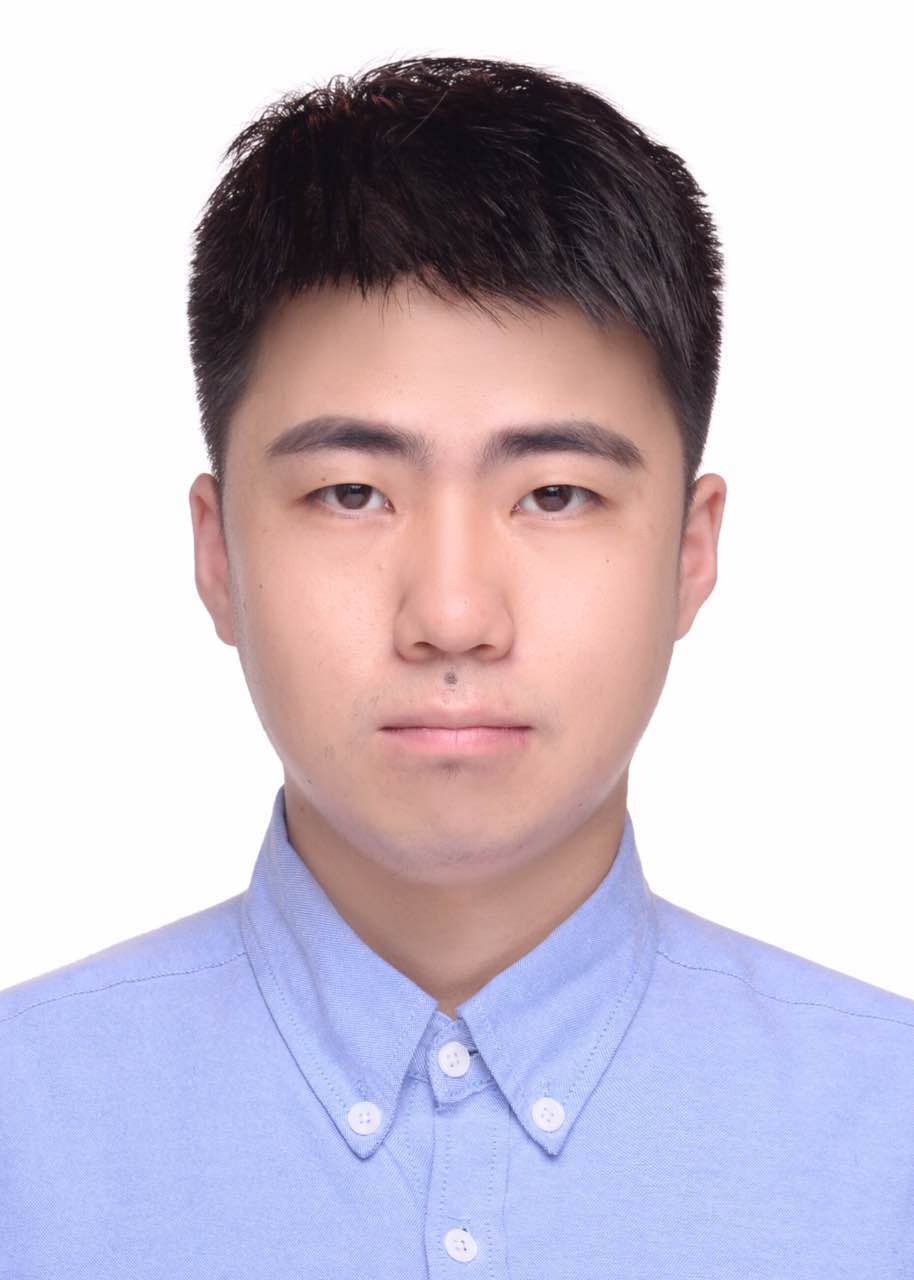}}]{Xiaogang Xu}  is currently a ZJU100 Professor in Zhejiang University. Meanwhile, he is a research scientist in Zhejiang Lab. Before that, he was a research fellow in the Chinese University of Hong Kong. He obtained his Ph.D. degree from the Chinese University of Hong Kong in 2022. He received his bachelor degree from Zhejiang University. He obtained the Hong Kong PhD Fellowship in 2018. He serves as a reviewer for CVPR, ICCV, ECCV, AAAI, ICLR, NIPS, IJCV. His research interest includes large generative models, multi-modality networks, etc.
\end{IEEEbiography}

\begin{IEEEbiography}[{\includegraphics[width=1in,height=1.3in,clip]{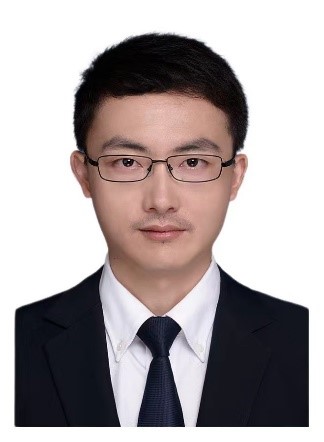}}]{Jiafei Wu} received the B.S. degree from JXUFE in 2010, the M.S. degree and Ph.D. degree from the University of Hong Kong in 2012 and 2017, respectively. He has been a senior engineer, manager and deputy director from 2018 to 2023 in SenseTime. He is currently with the Zhejiang Lab. His research interests include deep learning, trustworthy AI, embedded system, and computational intelligence.
\end{IEEEbiography}

\begin{IEEEbiography}[{\includegraphics[width=1in,height=1.3in,clip]{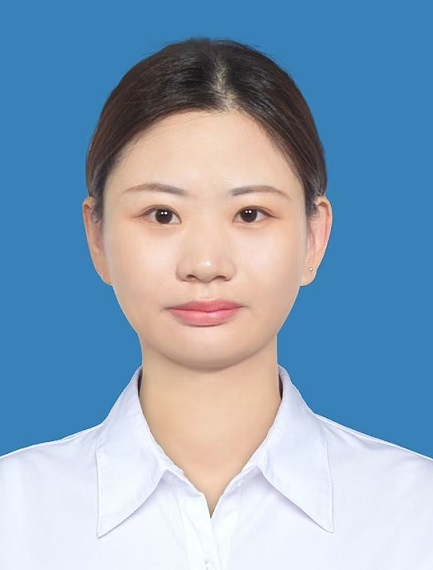}}]{Yan Han} received her bachelor degree from Huazhong Agricultural University, and received master degree from Wuhan University in 2019. She is currently working at Zhejiang Lab, where her research focuses on computer vision, image processing, and multimodal models,etc.
\end{IEEEbiography}

\begin{IEEEbiography}[{\includegraphics[width=1in,height=1.3in,clip]{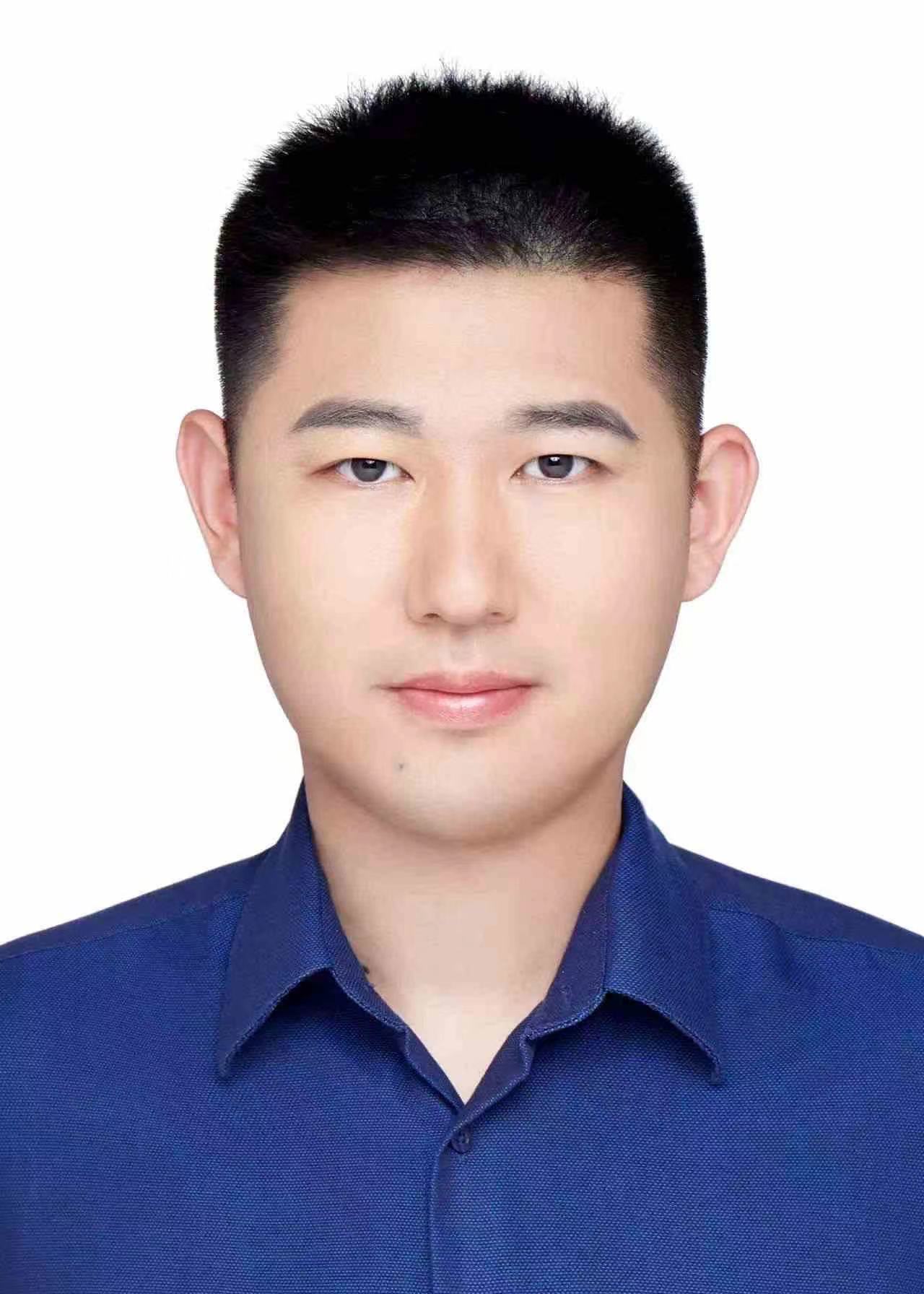}}]{Zhe Liu} received the BS and MS degrees from Shandong University, China, in 2008 and 2011, respectively, and the PhD degree from the Lab-oratory of Algorithmics, Cryptology and Security, University of Luxembourg, Luxembourg, in 2015. He is a professor with Zhejiang Lab. His research interests include security, privacy and
cryptography solutions for the Internet of Things.
\end{IEEEbiography}

\end{document}